\pdfoutput=1

\documentclass[letterpaper, 10pt,conference]{ieeeconf}

\IEEEoverridecommandlockouts

\usepackage{graphicx}
\usepackage{xcolor}
\usepackage{hyperref}
\usepackage{booktabs}
\usepackage{amssymb}
\usepackage{amsthm}
\usepackage{amsmath}
\usepackage{nomencl}
\usepackage{gensymb}
\usepackage{siunitx}
\usepackage{subfig}
\usepackage{dblfloatfix} 

\begin{document}

\title{RAM-VO: Less is more in Visual Odometry}

\author{
Iury Cleveston and Esther L. Colombini \\
Laboratory of Robotics and Cognitive Systems (LaRoCS) \\
Institute of Computing, University of Campinas \\
Campinas, São Paulo, Brazil \\
iury.cleveston@ic.unicamp.br,  esther@ic.unicamp.br
}

\maketitle

\begin{abstract}
Building vehicles capable of operating without human supervision requires the determination of the agent's pose. Visual Odometry (VO) algorithms estimate the egomotion using only visual changes from the input images. The most recent VO methods implement deep-learning techniques using convolutional neural networks (CNN) extensively, which add a substantial cost when dealing with high-resolution images. Furthermore, in VO tasks, more input data does not mean a better prediction; on the contrary, the architecture may filter out useless information. Therefore, the implementation of computationally efficient and lightweight architectures is essential. In this work, we propose the RAM-VO, an extension of the Recurrent Attention Model (RAM) for visual odometry tasks. RAM-VO improves the visual and temporal representation of information and implements the Proximal Policy Optimization (PPO) algorithm to learn robust policies. The results indicate that RAM-VO can perform regressions with six degrees of freedom from monocular input images using approximately 3 million parameters. In addition, experiments on the KITTI dataset demonstrate that RAM-VO achieves competitive results using only 5.7\% of the available visual information.
\end{abstract}


%
\IEEEpeerreviewmaketitle

\section{Introduction}
Autonomous vehicles have attracted significant attention in the last few years. These vehicles require a proper perception and understanding of the world to determine their localization. In these scenarios, Visual Odometry (VO) methods provide a solution by estimating the egomotion using only visual changes from the input images. These methods require the environment to have sufficient light, the objects to have texture, and the subsequent images to overlap. However, traditional VO methods still present severe issues in real-world environments due to sudden changes in the agent's speed, changes in the scene such as illumination, shadows, occlusions, and simultaneous motion of numerous objects~\cite{geiger_vision_2013}. In recent years, Deep Learning (DL) appeared as a novel way to learn, directly from the data, the various nonlinear factors that influence scene generation and motion~\cite{goodfellow_deep_2016}. DL methods commonly outperform either direct or indirect hand-crafted solutions and traditional learning methods that usually suffer from non-linearities~\cite{ciarfuglia_evaluation_2014,guizilini_semi-parametric_2013,roberts_memory-based_2008}.

However, deep learning methods for visual odometry make extensive use of convolutional neural networks (CNN), which add a substantial cost when dealing with high-resolution images. Further, more input data does not mean a better prediction; on the contrary, the network may have to learn how to filter out useless information. Therefore, the implementation of computationally efficient and lightweight architectures, especially for mobile devices, has attracted significant interest in approaching the problem from a new perspective. Though capturing only the necessary information is fundamental, learning where to look requires elaborating several cognitive concepts, such as attention. 

\begin{figure}[t!]
    \centering
    \subfloat{{\includegraphics[width=0.98\linewidth]{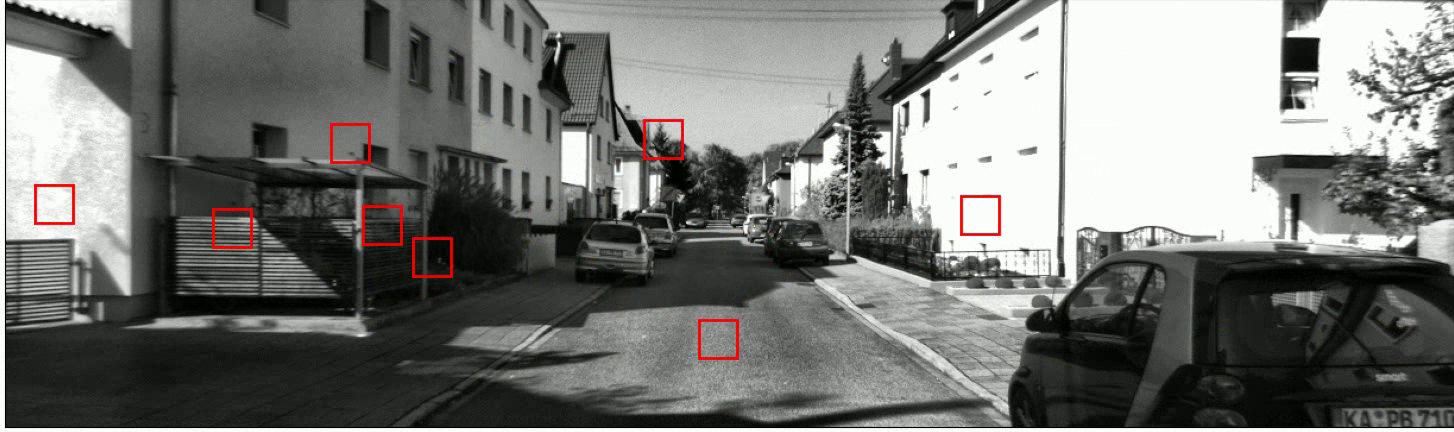} }}
    \vspace{-0.6em}
    \subfloat{{\includegraphics[width=0.48\linewidth]{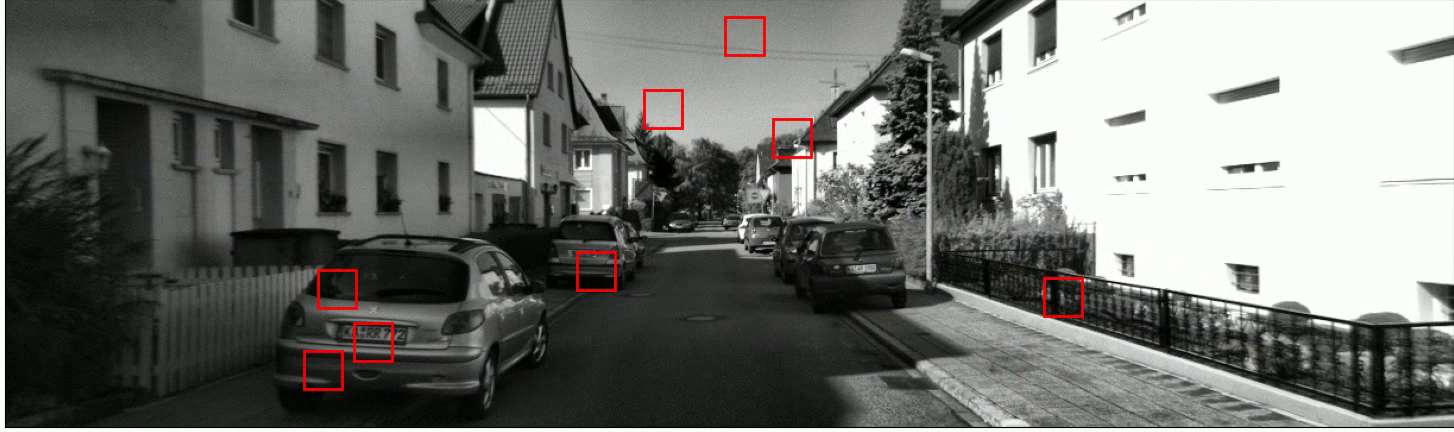} }}
    \subfloat{{\includegraphics[width=0.48\linewidth]{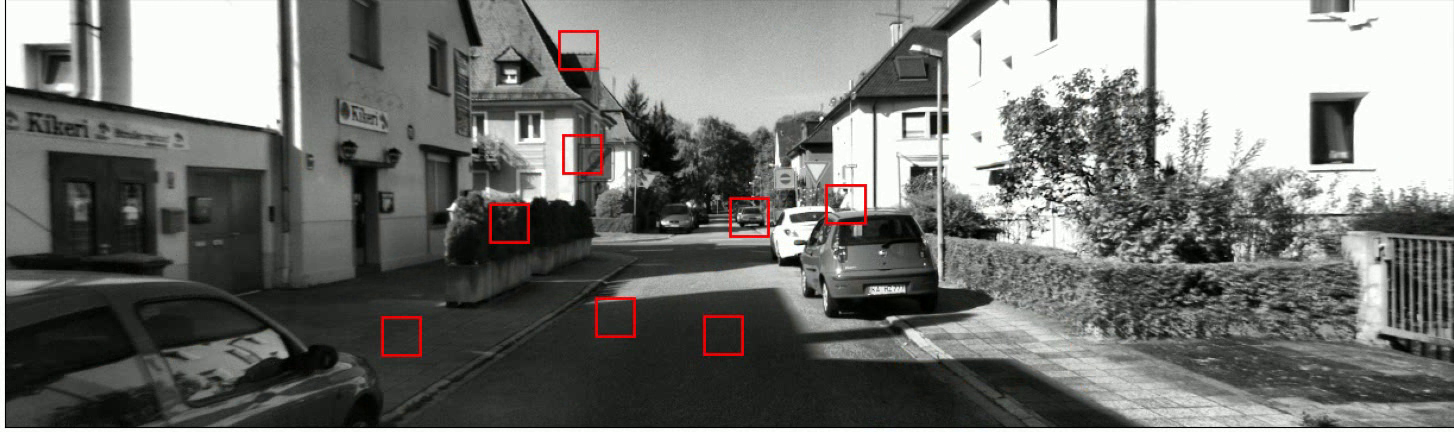} }}

    \caption{Sequences of eight observations by frame. A learned policy defines the patch location (red square). The total input information corresponds to only 5.7\% of the total available.}
    \label{fig:baseline_all_glimpses}
    \vspace{-1.5em}
\end{figure}

In this context, the Recurrent Attention Model (RAM) \cite{mnih2014recurrent} have emerged as a novel architecture, which implements a recurrent attentional glimpse, incorporating the attention concept by incrementally selecting the essential pieces of information. One of the RAM's main advantages is employing Reinforcement Learning (RL) to guide the glimpse sensor through the image; the RL paradigm allows the model to learn a more robust and efficient policy by trial and error. However, RAM was introduced mainly as a concept proof, only implemented for classification tasks on the MNIST~\cite{lecun1998mnist} dataset. RAM also uses the REINFORCE rule~\cite{williams1992simple} to guide the glimpse sensor, but this algorithm presents convergence problems and slowness in challenging scenarios.

Therefore, in this work, we propose a monocular end-to-end visual odometry architecture --- RAM-VO --- that employs the reinforcement learning paradigm to train an attentional glimpse sensor over time. The proposed RAM-VO architecture extends RAM by introducing spatial and temporal elements to enable the 6-degree-of-freedom (6-DoF) pose regression in real-world sequences. Furthermore, RAM-VO is more computationally efficient than similar VO methods due to the addition of attentional mechanisms and the use of Proximal Policy Optimization (PPO)~\cite{schulman2017proximal} to learn a robust policy. To the best of our knowledge, this is the first architecture to perform visual odometry that implements reinforcement learning in part of the pipeline. 

\subsection{Contributions}

This work provides the following contributions:

\begin{itemize}

    \item A lightweight VO method that selects the important input information via attentional mechanisms;
    
    \item The first visual odometry architecture that implements reinforcement learning in part of the pipeline;
    
    \item Several experiments on KITTI~\cite{geiger_vision_2013} sequences demonstrating the validity and efficiency of RAM-VO.

\end{itemize}


\section{Related Work}

The visual odometry field has seen a massive increase in architectures and publications in recent years. In 2015, Konda et al.~\cite{konda_learning_2015} proposed the first architecture in the field, implemented as an end-to-end CNN model to estimate direction and velocity from raw stereo images. In the same period, extracting the optical flow was a common practice to initialize the models~\cite{muller_flowdometry_2017}. Subsequent architectures coupled LSTM layers to provide temporal representation~\cite{mohanty_deepvo_2016}. In 2017, Wang et al.~\cite{wang_deepvo_2017} proposed DeepVO, which is an end-to-end monocular architecture capable of extracting visual features directly from the input raw images with a CNN; and determining temporal relation with an LSTM. 

After that, several supervised learning methods appeared to tackle distinct issues in the field. Peretroukhin and Kelly~\cite{peretroukhin_dpc-net_2018} proposed the DPC-Net architecture, which aims to integrate the representation capabilities of deep neural networks with the efficiency of geometric and probabilistic algorithms. DPC-Net implements a CNN-based architecture to learn the corrections for the pose estimator. Zhao et al.~\cite{zhao_learning_2018} propose the L-VO architecture, which predicts the 6-DoF pose from 3D optical flow for monocular VO. Valada et al.~\cite{valada_deep_2018} proposed the VLocNet architecture, which is capable of estimating the 6-DoF pose in a monocular setup. Their architecture fuses relative and global RCNN-based architectures to improve accuracy. Saputra et al.~\cite{saputra_distilling_2019} proposed to distill knowledge from a pose regressor employing concepts like Knowledge Distillation (KD) to train two networks jointly. Saputra et al. also proposed the first Curriculum Learning (CL) architecture, called CL-VO~\cite{saputra_learning_2019}, aiming to learn the scene geometry for monocular VO by gradually increasing the task's difficulty.

Several methods started to use concepts of attention in their pipeline in recent years, mainly for unsupervised learning. In 2020, Damirchi et al.~\cite{damirchi2020exploring} explored the concept of self-attention to extract meaningful features in complex scenarios, which usually have many moving objects and low textures.  Also, Kuo et al.~\cite{kuo2020dynamic} proposed the DAVO architecture, which is a dynamic attention-based visual odometry composed of two attentional networks. Their first network can generate semantic masks for determining the weights that each piece of the input image should have, while the second network uses a squeeze-and-excite attentional block. 

Attentional concepts are also employed to support the salient feature extraction from the input images. Liang et al.~\cite{liang_salientdso_2018} proposes the SalientDSO architecture, which applies attention in the Direct Sparse Odometry (DSO)~\cite{engel_direct_2018} algorithm. The proposed method runs in two distinct modules, the first one detects the visual salience using SalGAN, and the second one performs visual odometry with DSO. The drift error has substantially decreased compared to the original DSO due to the improved sampling. Chen et al.~\cite{chen_salient_2019} also proposed an end-to-end CNN+LSTM salient-feature attention and context-guided networks for robust visual odometry. Their architecture can be trained using only monocular images and aimed to decouple rotational and translational motion.

Deep learning methods demand the adoption of large and representative datasets; conversely, lightweight and efficient methods are fundamental to the field. Reinforcement Learning (RL) and attention applied to visual odometry can provide a solution in such scenarios. The architecture becomes efficient by selecting only the necessary input data, and learning a robust policy can mitigate the drift error. However, we have not found any method that implements RL in any part of the visual odometry pipeline.

\subsection{The Recurrent Attention Model (RAM)}

\begin{figure}[ht]
    \centering
    \includegraphics[width=1\linewidth]{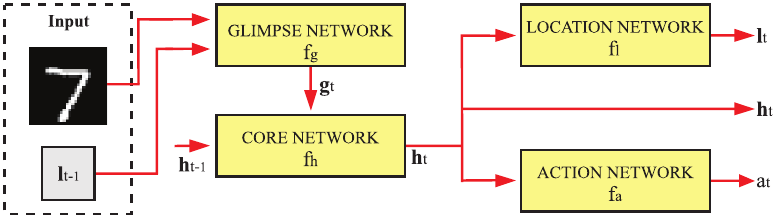}
    \caption{The RAM architecture is composed of four distinct networks, each one encapsulating a specific behavior.}
    \label{fig:ram}
\end{figure}

The Recurrent Attention Model (RAM)\cite{mnih2014recurrent} implements a hard attention mechanism similar to the biological visual system, which iteratively builds an informative vector through multiple observations in the input image. Hard attention forces the model to consider only the relevant elements, discarding the others entirely~\cite{correia2021attention}.
The observations are iteratively stored in a latent space, providing the knowledge to perform the task. The location of each observation is determined by a policy learned through REINFORCE~\cite{williams1992simple}. RAM is composed of four distinct networks (Figure~\ref{fig:ram}).

\begin{figure*}[bp]
    \centering
    \includegraphics[width=0.99\linewidth]{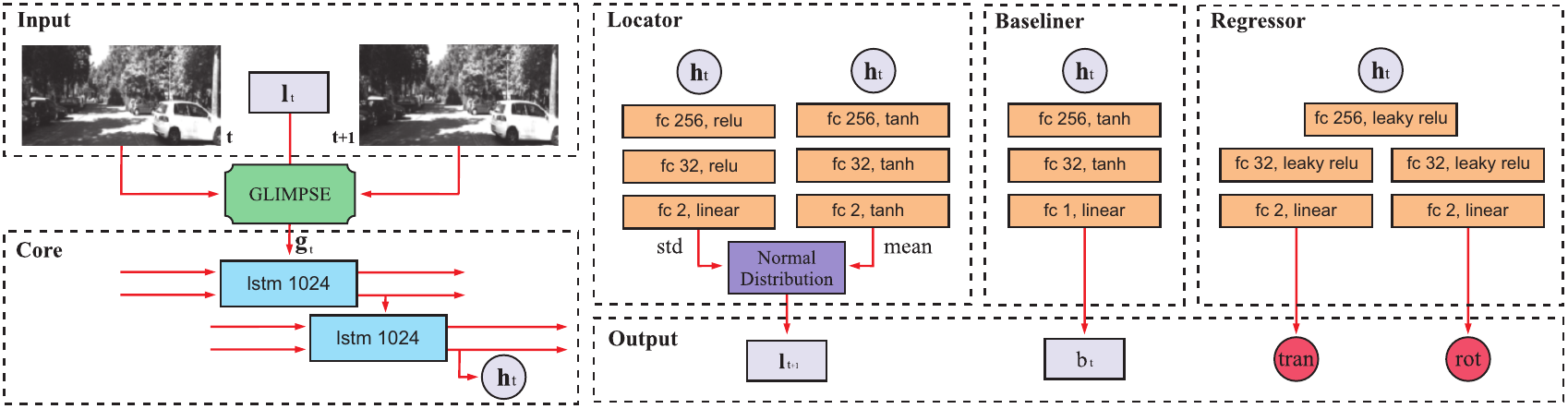}
    \caption{RAM-VO is composed of five subnetworks that perform specific functions. The Glimpse allows efficient data consumption by successively observing regions of interest $\textbf{l}_t$ on the input images. The Core sequentially integrates these observations $\textbf{g}_t$ in nested LSTMs, the Locator generates the next observation's location $\textbf{l}_{t+1}$ by sampling a Gaussian distribution parametrized by the internal state $\textbf{h}_t$, and the Regressor predicts the 6-DoF pose regression in the end.}
    \label{fig:ramvo_baseline}
\end{figure*}

The glimpse network $f_{g}$ represents the attentional system and comprises a glimpse sensor to extract meaningful patches from the input images, and fully connected layers to encode the visual information. First, the glimpse sensor receives an image $\textbf{x}_{t}$ and a location $\textbf{l}_ {t-1}$ as input. Then, several patches are extracted in different resolutions, centered at the location $\textbf{l}_{t-1}$. This process builds a pyramidal-like structure $\rho (\textbf{x}_{t}, \textbf{l}_{t-1})$ similar to biological vision, representing what was observed on the image. Finally, the glimpse network concatenates $\rho_{t}$ and $\textbf{l}_{t-1}$ to include the location where the information was extracted, resulting in the final vector $\textbf{g}_{t}$.

The core network $f_{h}$ stores the multiple observations by receiving the glimpse feature vector $\textbf{g}_{t}$ and the previous internal state $\textbf{h}_{t-1}$ as input at every time step $t$. Through fully connected layers, the core network outputs the current internal state $\textbf{h}_{t}$, which condenses all the sequential information provided by the glimpse network. The location network $f_{l}$ generates the location $\textbf{l}_t$ for the subsequent observation by sampling a Gaussian distribution with two dimensions $(x,~y)$ and a fixed standard deviation. For each subsequent observation, a novel Gaussian distribution is generated by using the internal state $\textbf{h}_{t}$ to parameterize the mean $\mu_t$. After all observations, the action network $f_{a}$ consumes the internal state $\textbf{h}_{t}$ to predict the class $a_{t}$, which is the ultimate goal.

\begin{figure*}
    \centering
    \includegraphics[width=0.99\linewidth]{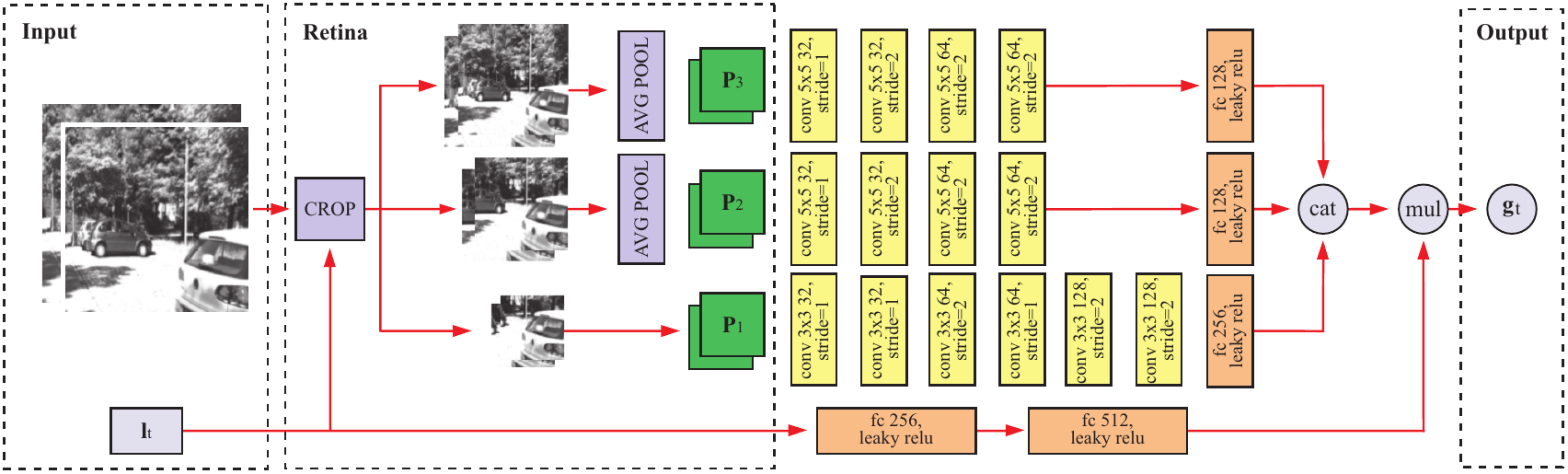}
    \caption{The Glimpse Network. The consecutive input images are concatenated and cropped at the location $\textbf{l}_t$, generating three patches $\textbf{P}_{1,2,3}$. The patches are processed by independent CNN pipelines and concatenated in an internal vector; further multiplied by the encoded location $\textbf{l}_t$, providing the final glimpse vector $\textbf{g}_t$. This design favors the learning of optical flow.}
    \label{fig:ramvo_glimpse_baseline}
    \vspace{-1em}
\end{figure*}

The hard attention mechanism requires reinforcement learning strategies to train the model. Therefore, the RL setup is an instance of a Partially Observable Markov Decision Process (POMDP), in which the true state of the environment is unobserved; hence the model needs to learn a stochastic policy  $\pi \left (\textbf{l}_{t} | \textbf{h}_{1:T};\boldsymbol\theta \right )$ that maps the environment history $\textbf{h}_{1:T} = \{x_{1}, l_{1}, ..., x_{t}, l_{t}\} $ to a distribution over the actions in time step $t$, restricted by the sensor. In this sense, the glimpse sensor is the agent, the whole image is the environment, and the rewards are defined according to the success in the classification. From the ground-truth values, the agent receives $r_{t} = 1$ if the class is classified correctly after $T$ time steps, and 0 otherwise. The goal is to maximize the return $G = \sum_{t=1}^{T}r_{t}$, which is sparse and delayed. 

The architecture parameters $\boldsymbol\theta$ are optimized by maximizing the return $G$ when the agent interacts with the environment. The agent's policy, in combination with the environment's dynamics, produces a distribution over the possible iteration sequences $\textbf{h}_{1:N} $, and the goal is to maximize the return under that distribution via $J(\boldsymbol\theta) = \mathbb{E}_{p(\textbf{h}_{1:T};\boldsymbol\theta)}[G]$, where $p(\textbf{h}_{1:T};\boldsymbol\theta)$ depends on the policy. Maximizing $J(\boldsymbol\theta)$ is not trivial because it involves an expectation about high-dimension iteration sequences, which may involve unknown environment dynamics. However, we can obtain an approximation of the gradient with the REINFORCE rule~\cite{williams1992simple} as

\begin{equation}
\nabla J(\boldsymbol\theta) = \frac{1}{M}\sum_{i=1}^{M}\sum_{t=1}^{T} \nabla \mathrm{log}\pi\left ( \textbf{l}_{t}^{i} | \textbf{h}_{1:t}^{i}; \boldsymbol\theta\right )(G^{i}_{t} - b_{t}),
\end{equation}

\noindent where $\textbf{h}^{i}_{1:t}$ are the sequences obtained by running the current agent policy $\pi_{\theta}$ for $i=1, ..., M$ episodes, $G^{i}_{t}$ is the accumulated reward obtained after executing action $\textbf{l}_t^i$, and $b_{t}$ is the baseline value, which reduces the variance for the gradient updates. The baseline value $b_{t}$ depends on sequence $\textbf{h}^{i}_{1:t}$ but not directly of the action $\textbf{l}^{i}_{t}$. As a result, the algorithm increases the log-probability of actions that generate a high cumulative reward and diminishes the probability of actions that generates a low cumulative reward.

RAM is a hybrid architecture in which the location network is trained by the REINFORCE rule~\cite{williams1992simple}, while the other networks employ supervised learning. Although bringing innovative ideas from biology, RAM was proposed mainly as a proof of concept, lacking the necessary complexity to deal with high-resolution images and regression tasks.

\section{The RAM-VO Architecture}

The Recurrent Attentional Model for Visual Odometry (RAM-VO) was constructed by extending RAM~\cite{mnih2014recurrent} and changing the architecture's goal from classification to regression. We propose several modifications to learn a robust policy, deal with complex visual information, and enable 6-DoF pose regressions --- the next sections detail each modification. RAM-VO is shown in Figures~\ref{fig:ramvo_baseline} and \ref{fig:ramvo_glimpse_baseline}.

\subsection{Glimpse Network}

\begin{figure}[b!]

    \centering
     \includegraphics[width=1\linewidth]{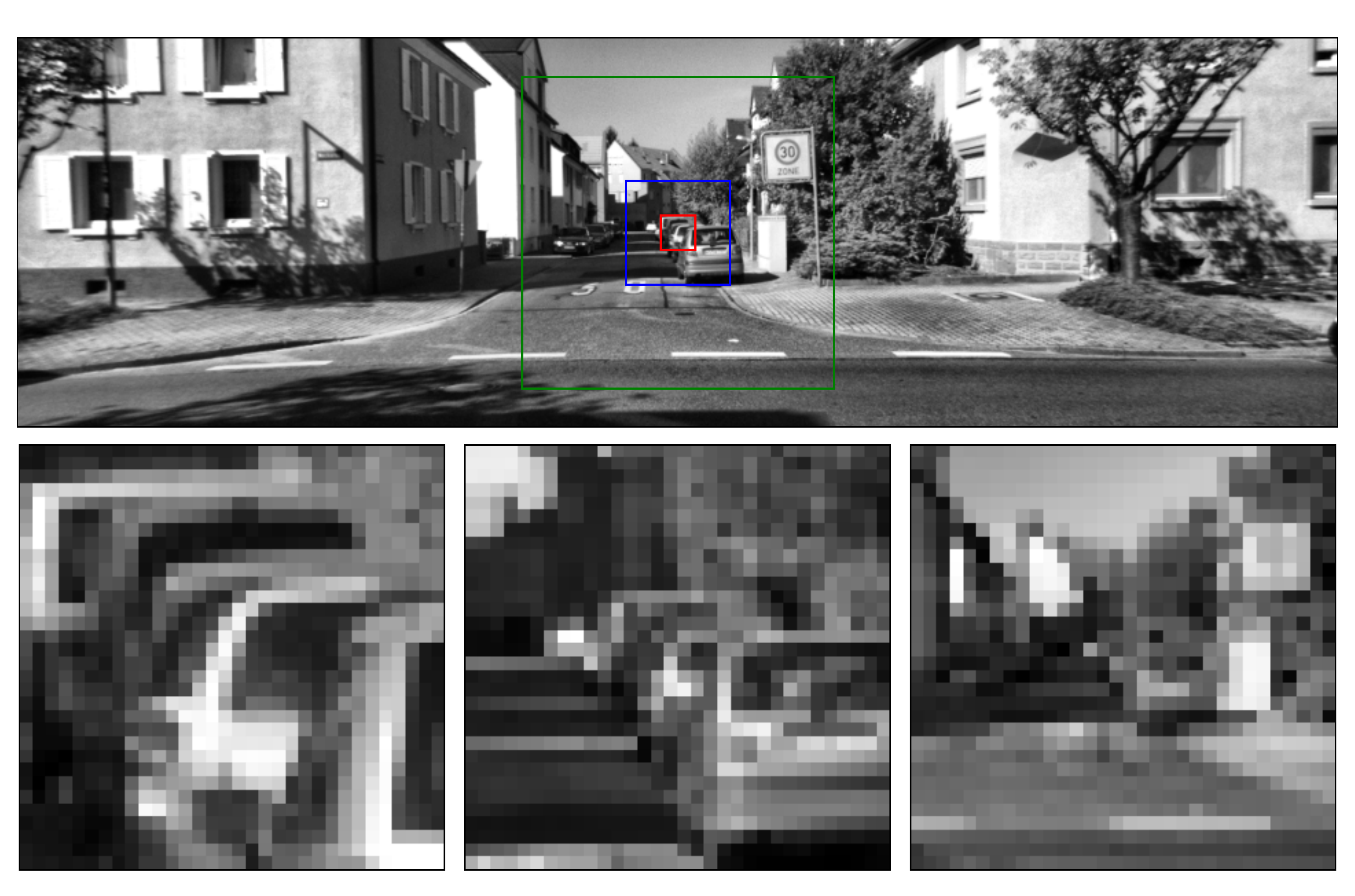}

    \caption{The glimpse sensor extracts three patches from the input image at a given location (top row). Each patch has an increased size, forming a pyramidal-like structure. The first patch $\textbf{P}_{1}$ is cropped with $32 \times 32$ pixels; the others $\textbf{P}_{2,3}$ are twice larger than the previous one (bottom row).}
    \label{fig:city_scales}
\end{figure}

The visual odometry task requires processing two consecutive frames in the glimpse network, allowing the detection of the same features in both images and their correspondence. In this sense, the glimpse network receives two temporally consecutive images $(\textbf{x}_t, \textbf{x}_{t+1})$ and a location of interest $\textbf{l}_t$ as input. The images are stacked and cropped at the location provided, generating three patches of 32, 64, 128 pixels. An average pooling operation is performed in the larger patches to reduce their size to $32 \times 32$ pixels. Then, the three final patches $\textbf{P}_{1,2,3}$ are processed through convolutional layers, flattened, and concatenated into a temporary internal vector. The patches locations are encoded and multiplied by the internal vector, originating the final glimpse vector $\textbf{g}_t$. In this sense, the glimpse network implements a top-down attentional mechanism by capturing small portions of visual information that we will fuse with the information locations to generate a final vector with the correspondence between the two input images. 

Generating an efficient representation $\textbf{g}_t$ for the input images is closely associated with learning the geometry presented in the scene. Therefore, based on the FlowNetS~\cite{fischer_flownet_2015}, we proposed to apply the convolutional operations in both images simultaneously to learn the optical flow and considerably reducing the computational cost compared to extracting the features separately. In this sense, the glimpse network uses 6 CNN layers with 32 to 128 channels for the patch $\textbf{P}_1$, and 4 CNN layers with 32 to 64 channels for the other two patches $\textbf{P}_{2,3}$. We observed that convolutional operations with smaller kernels significantly improve the representation; therefore, we processed the patch $\textbf{P}_1$ with a kernel size of $3 \times 3$ pixels and the others with a $5 \times 5$ kernel. Padding values are defined as zero to avoid reducing the input size across the CNN layers, and we remove pooling operations entirely. We perform dimensionality reduction by varying the kernel stride between 1 and 2.

\textbf{Glimpse Scales}. The extraction of the image patches is performed in three different scales, simulating the human visual system \cite{mnih2014recurrent}. The first scale corresponds to a central, high-resolution region but with a smaller dimension, and the second and third scales present larger dimensions but lower resolutions (Figure~\ref{fig:city_scales}). This pyramidal-like structure provides a trade-off between the amount of information and computational cost. Therefore, the agent can observe the environment's details on the center and also the most salient elements presented on the boundaries. Also, the peripheral information helps the agent to determine the next location of interest $\textbf{l}_{t+1}$ for subsequent observations. 

\subsection{Core Network}

The core network is responsible for integrating all observations from the input images and providing the internal state $\textbf{h}_t$ for the pose regression and for generating the next location of interest $\textbf{l}_{t+1}$. Therefore, the observation $\textbf{g}_t$ received from the glimpse network is recurrently integrated, updating the internal state $\textbf{h}_t$ for every step. This process is repeated accordingly to the number of steps previously defined. In this sense, to track long-term dependencies and better represent the internal state, we adapted the core network to include two stacked LSTM layers with 1024 hidden units. LSTMs also diminished the problem of vanishing gradients during training, stabilizing the model. The ability to generate efficient RL policies is directly dependent upon the representation of the internal state $\textbf{h}_t$, especially during the first epochs when high exploration is desired; hence, we initialized the weights orthogonally for $\textbf{h}_t$.

\subsection{Locator and Baseliner Network}

The locator network provides the next location of interest $\textbf{l}_{t+1}$ by sampling a Gaussian distribution, whose mean and standard deviation are parametrized by the internal state $\textbf{h}_t$. The first location is defined randomly, and the others are sampled accordingly to the learned policy. Unlike the original RAM, the policy's standard deviation is also learned during training, promoting exploration in the first epochs. The locator network is detached from the supervised graph and trained separately by the REINFORCE rule~\cite{williams1992simple}. We jointly train a baseliner network to provide the state value $b_t$ for each step, reducing the variance between the returns. Both locator and baseliner networks are composed of fully connected layers with 256 to 32 hidden units.

\subsection{Regressor Network}

The regressor network generates the pose prediction after the last observation is integrated into the internal state $\textbf{h}_t$. The predictions are decoupled in rotational and translational components, which comprehends the orientation $\boldsymbol{\varphi}$ with the Euler angles as roll $\phi$, pitch $\theta$, and yaw $\psi$; and the position $\boldsymbol{p}$, composed of the coordinates $x$, $y$, and $z$. The main goal is to regress the 6-DoF pose vector $[\phi, \theta, \psi, x, y, z]^T$ for every pair of frames. In this sense, the regressor network comprises three fully connected layers from 256 to 32 hidden units; the last layers provide linear outputs for the prediction.

\subsection{Loss and Reward Function}

RAM-VO is a hybrid architecture where the regressor, core, glimpse, and baseliner networks are trained in a supervised learning fashion, while the locator is trained by reinforcement learning. The supervised loss and the reward function are both defined in terms of the MSE to minimize the outliers as much as possible since only one poor prediction can harm the entire trajectory. Therefore, the supervised loss $L$ is defined as

\begin{equation}
L = \frac{1}{N}\sum_{i=1}^{N} \left\| \boldsymbol{\hat p} - \boldsymbol{p} \right\|_2^2 + k \left\| \boldsymbol{\hat \varphi} - \boldsymbol{\varphi} \right\|^2_2,
\label{eq:loss_ramvo_supervised}
\end{equation}

\noindent where $\boldsymbol{\hat p}$ and $\boldsymbol{\hat \varphi}$ are the position and orientation prediction, respectively; $\boldsymbol{p}$ and $\boldsymbol{\varphi}$ are the ground-truth values; and $k$ is the constant factor weighting the two losses, favoring the rotational or the translational component; we kept the ground-truth values normalized and $k=1$. The reward function $R$ is defined as 

\begin{equation}
R = \frac{1}{1 + L}.
\label{eq:loss_ramvo_reward}
\end{equation}

We prefer not to bias the RL agent towards a specific behavior; therefore, only the visual odometry error $L$ is employed in the reward function. 

\subsection{Proximal Policy Optimization (PPO)}

The REINFORCE rule~\cite{williams1992simple} is known for presenting converge issues and slowness; this occurs by sudden updates on the policy's parameters, which can harm the entire training by converging to suboptimal solutions. Proximal Policy Optimization (PPO)~\cite{schulman2017proximal} aims to attenuate these problems by updating the policy inside trusted regions. Therefore, PPO's surrogate function determines that the current policy must be close to the last one, avoiding large parameters shift. In practice, the PPO implementation consisted of replacing the locator and baseliner network with a similar structure in terms of layers and hidden units. We also use memory replay to enable the policy refinement with already sampled data, improving the architecture's efficiency. The policy refinement proportion is 20:1 concerning the supervised network --- we want the best policy to control the input information flow.

\section{Materials and Methods}

\subsection{Dataset}

In this work, we used the KITTI dataset~\cite{geiger_vision_2013}, one of the most popular datasets for evaluating visual odometry methods. The entire dataset consists of 22 sequences (39.2 km) of real-world traffic data captured by a car moving across urban and rural areas in Germany. However, only the first eleven (00-10) sequences have ground-truth information. Therefore, we used the grayscale images provided by the left camera, resized to $1200 \times 360$ pixels in resolution. 

In order to compare our results with other methods, we chose sequences 0, 2, 4, 5, 6, 8, 9 for training, sequences 10 for validation, and sequences 3, 7 for testing. We did not use sequence 1 due to the higher average variation for the translational component. The whole data used comprehends 18,990 images for training, 1,200 for validation, and 1,902 for testing. We did not use data augmentation.

We preprocessed each image by equalizing the pixel intensity histogram in small windows of $8 \times 8$ pixels with the Contrast Limited Adaptive Histogram Equalization method. In this way, we can highlight the image's features without increasing noise. Also, we normalized the images with the z-score function before entering the glimpse network.

\begin{figure*}
    \centering
    \subfloat{{\includegraphics[width=4.3cm]{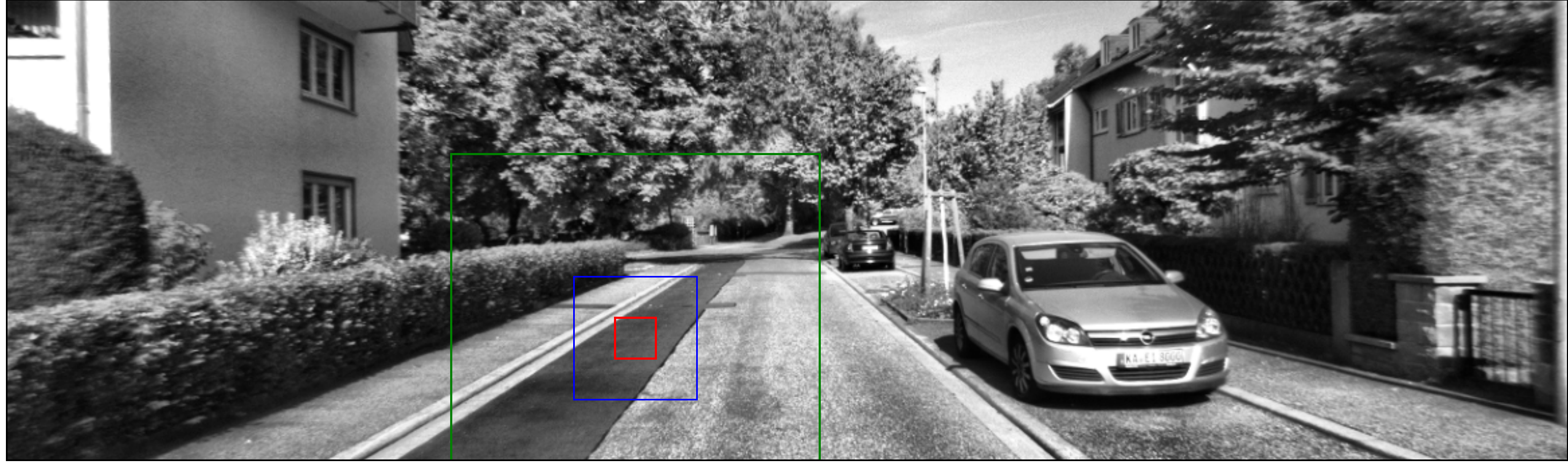} }}\hfill
    \subfloat{{\includegraphics[width=4.3cm]{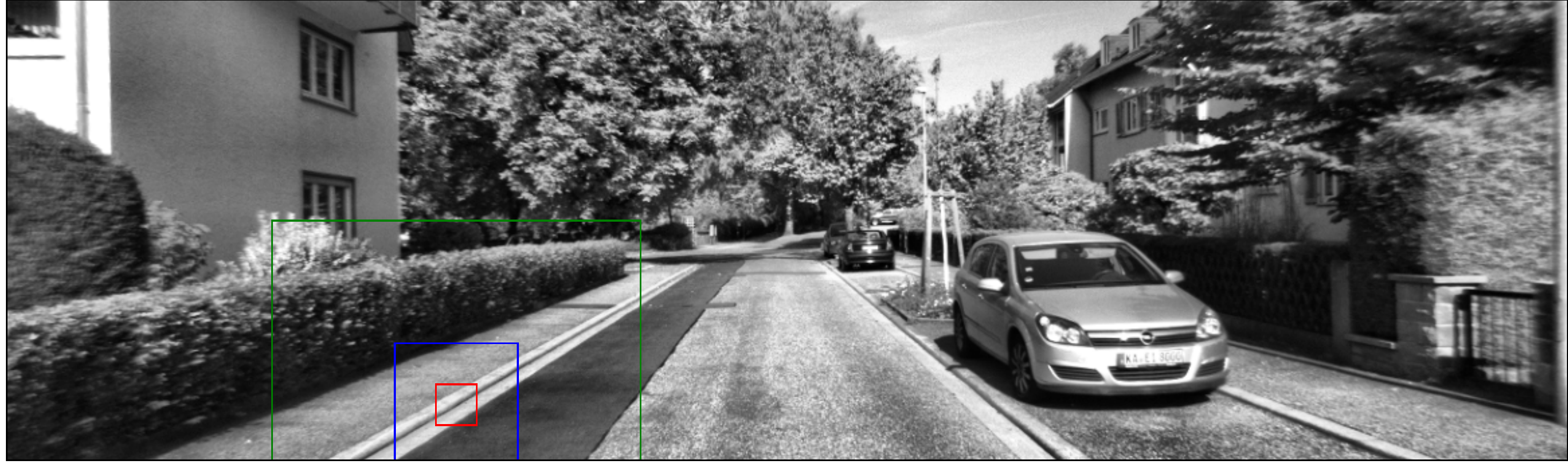} }}\hfill
    \subfloat{{\includegraphics[width=4.3cm]{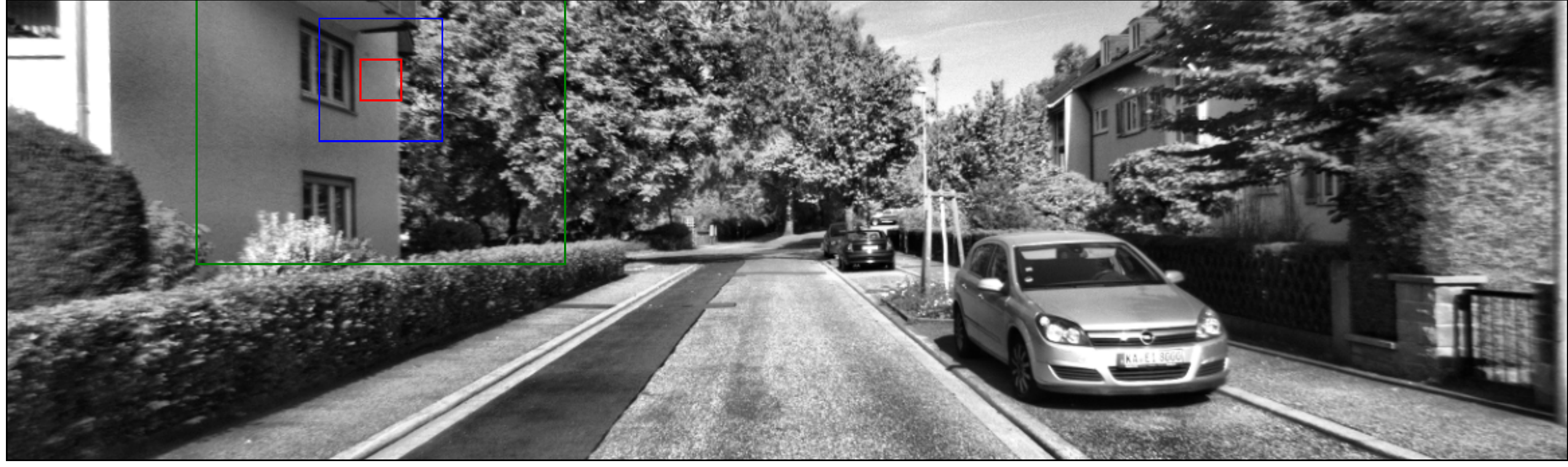} }}\hfill
    \subfloat{{\includegraphics[width=4.3cm]{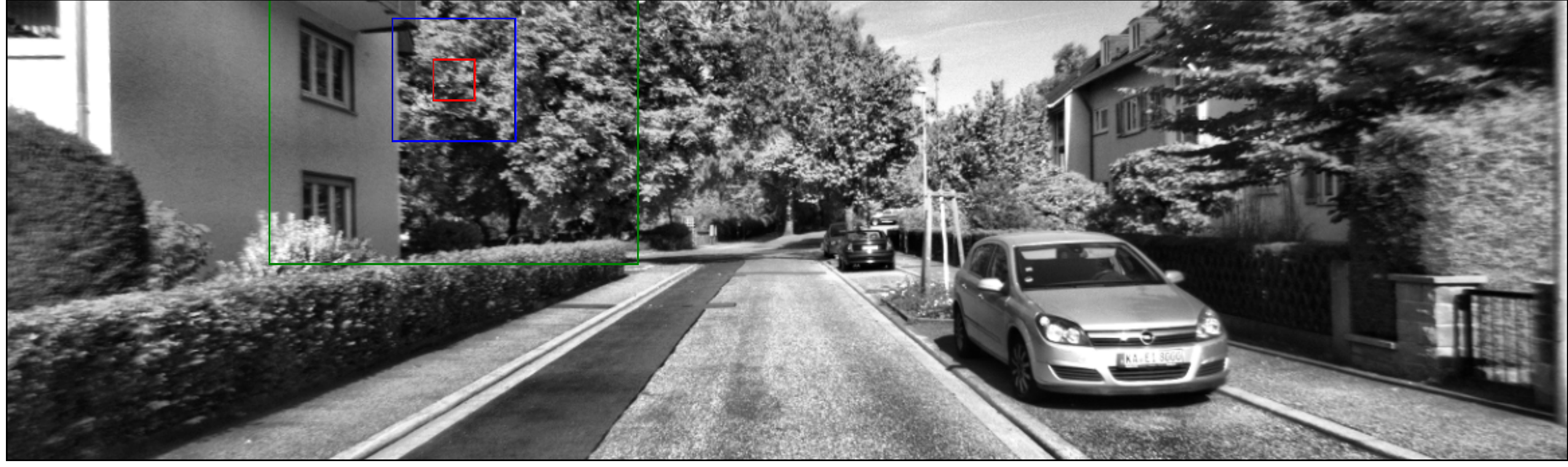} }}\hfill
    \vspace{-0.6em}
    \subfloat{{\includegraphics[width=4.3cm]{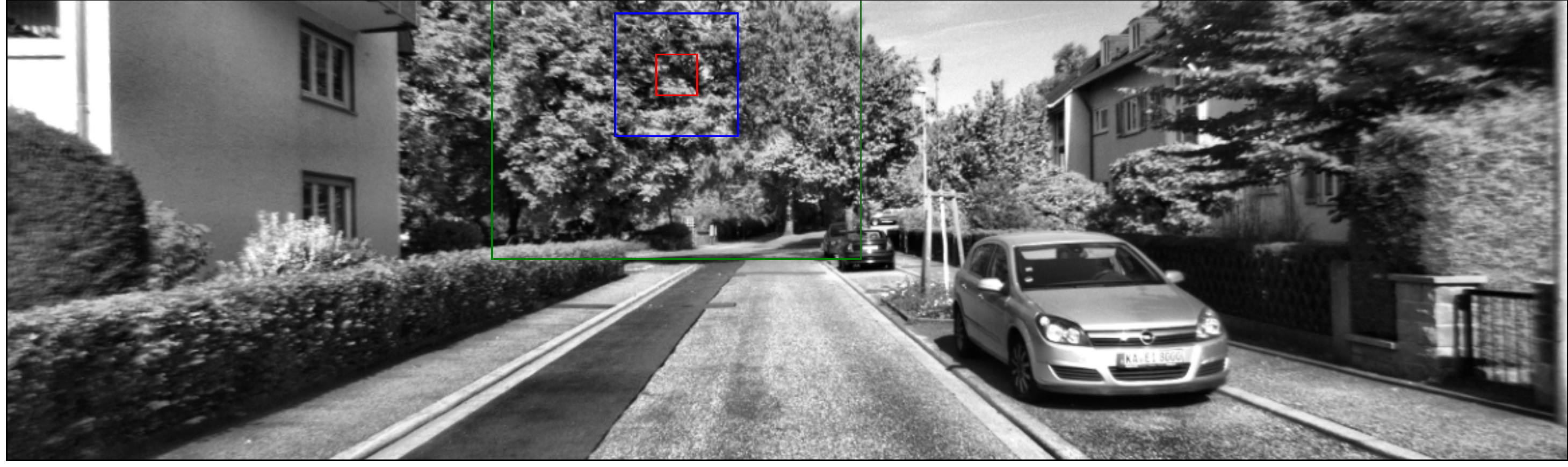} }}\hfill
    \subfloat{{\includegraphics[width=4.3cm]{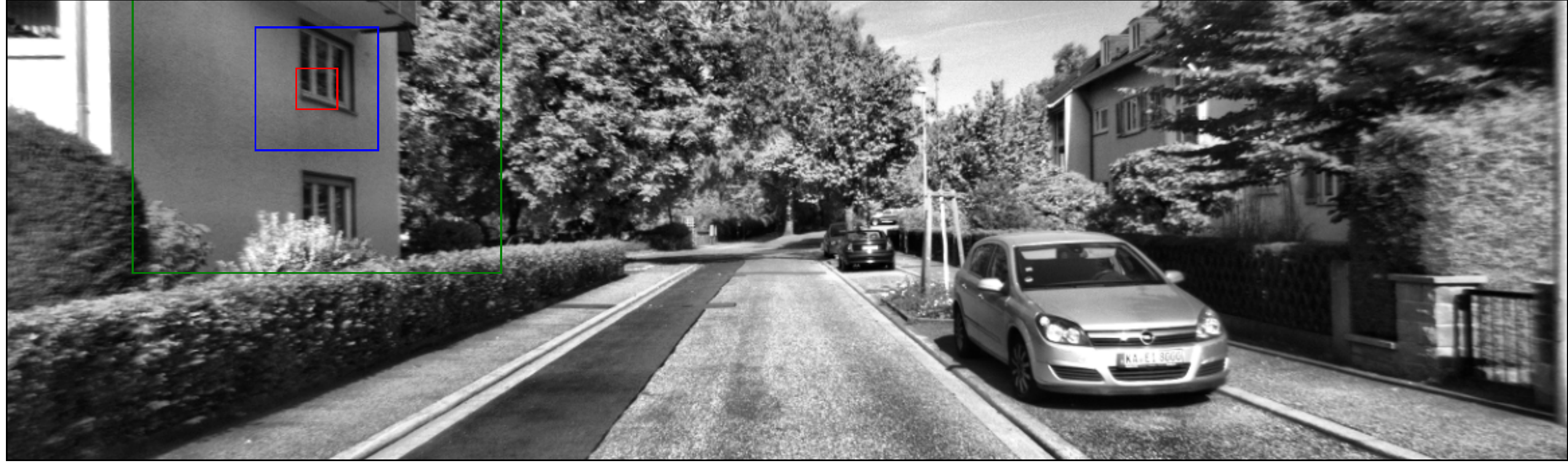} }}\hfill
    \subfloat{{\includegraphics[width=4.3cm]{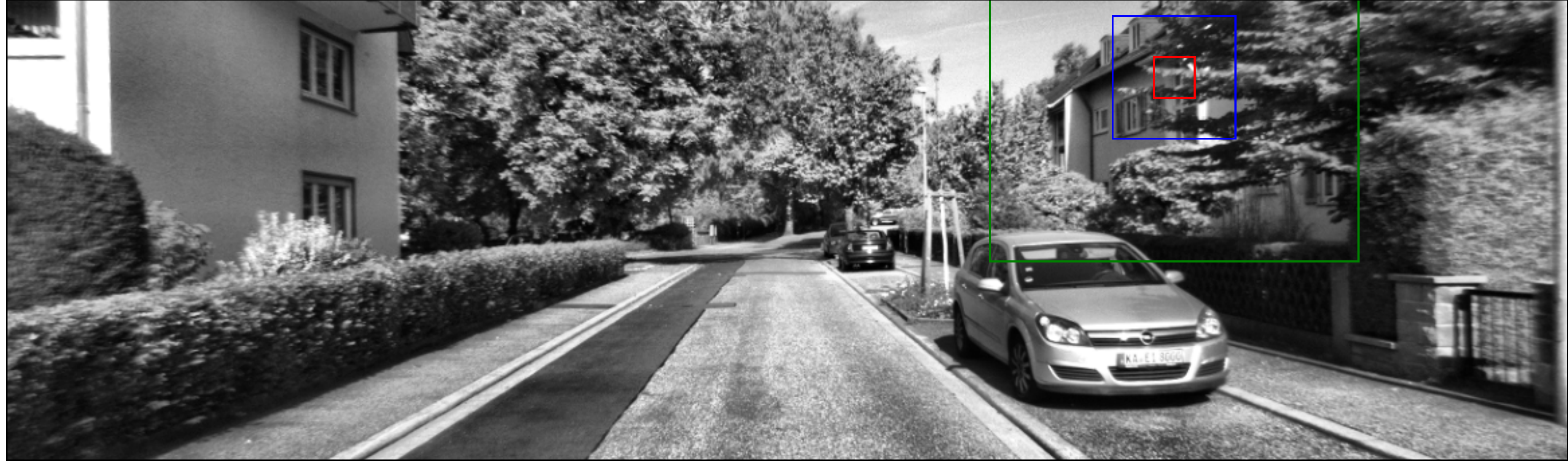} }}\hfill
    \subfloat{{\includegraphics[width=4.3cm]{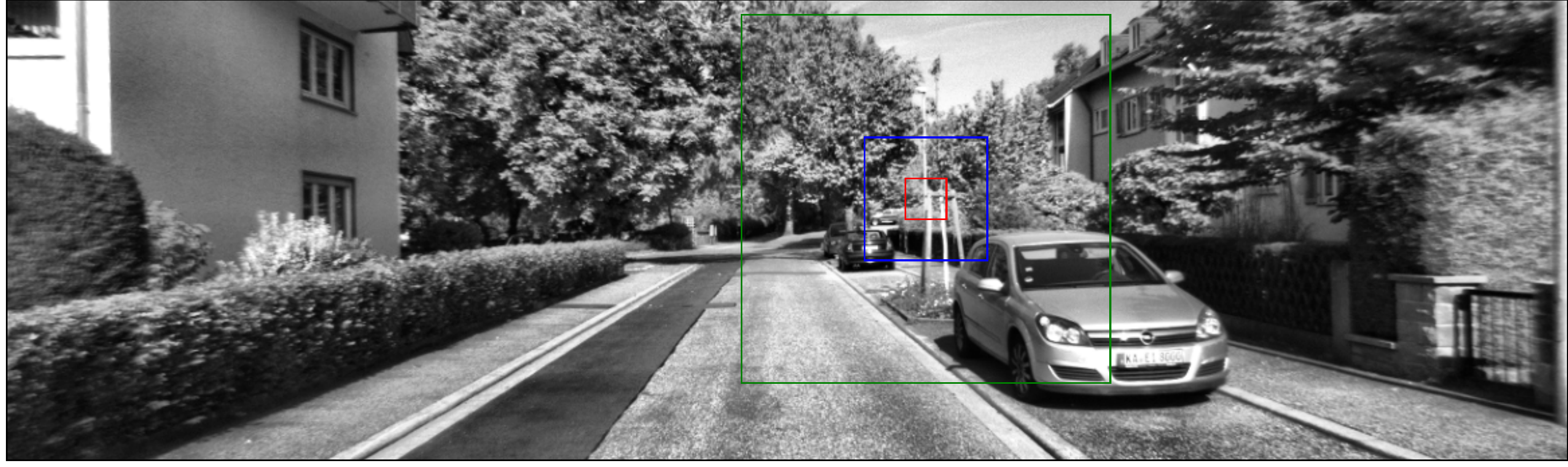} }}\hfill
    
    \caption{The temporal sequence of eight observations on the first frame on sequence 2. The glimpse sensor exploits high-gradient regions in most captures, such as corners and edges.}
    \label{fig:baseline_glimpses}
    \vspace{-1.5em}
\end{figure*}

\begin{figure*}
    \centering
    \subfloat{{\includegraphics[width=4.3cm]{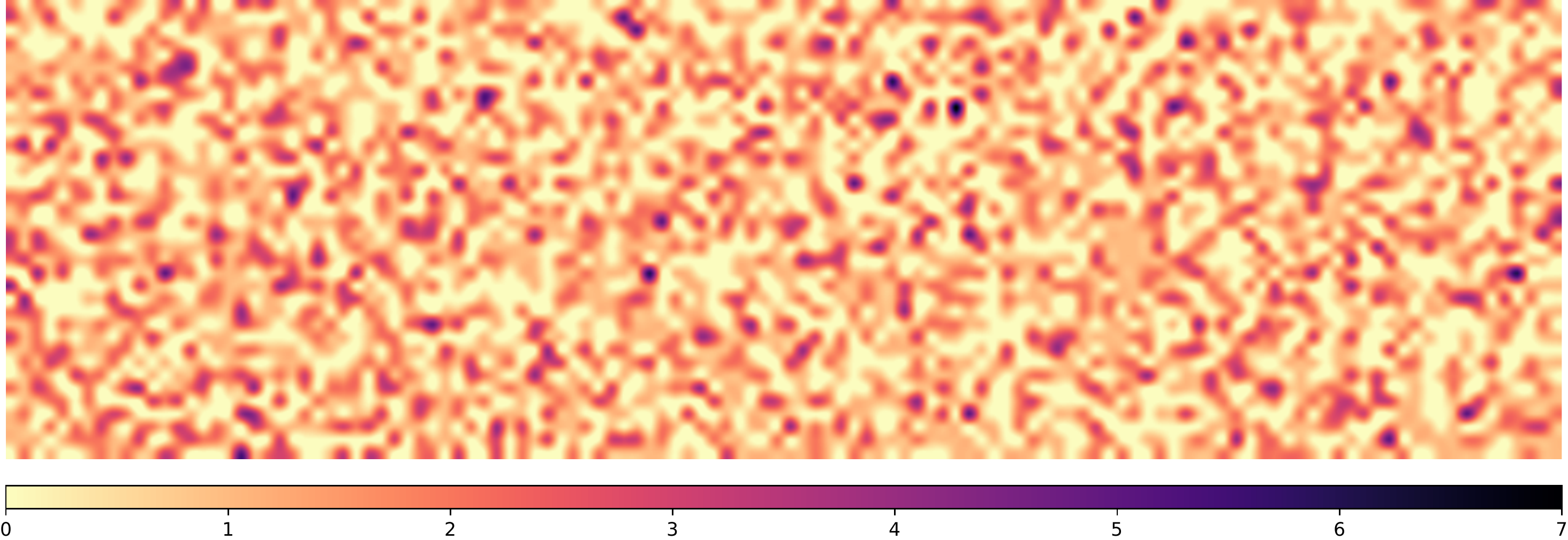} }}
    \subfloat{{\includegraphics[width=4.3cm]{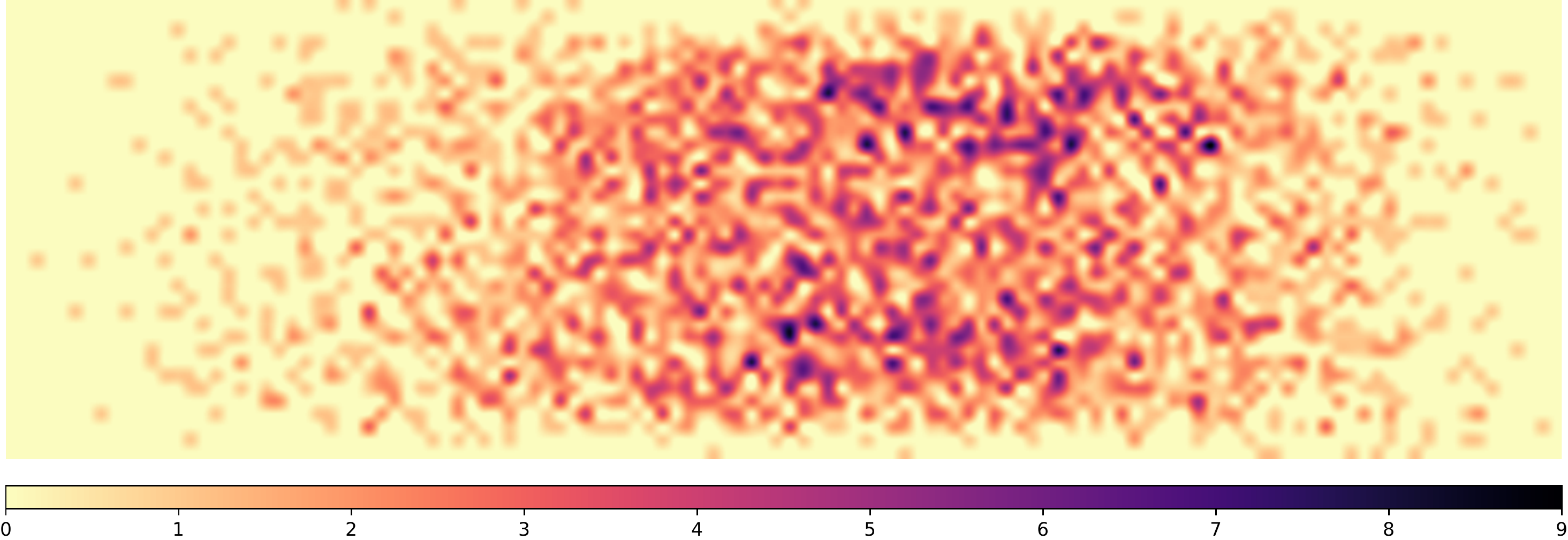} }}
    \subfloat{{\includegraphics[width=4.3cm]{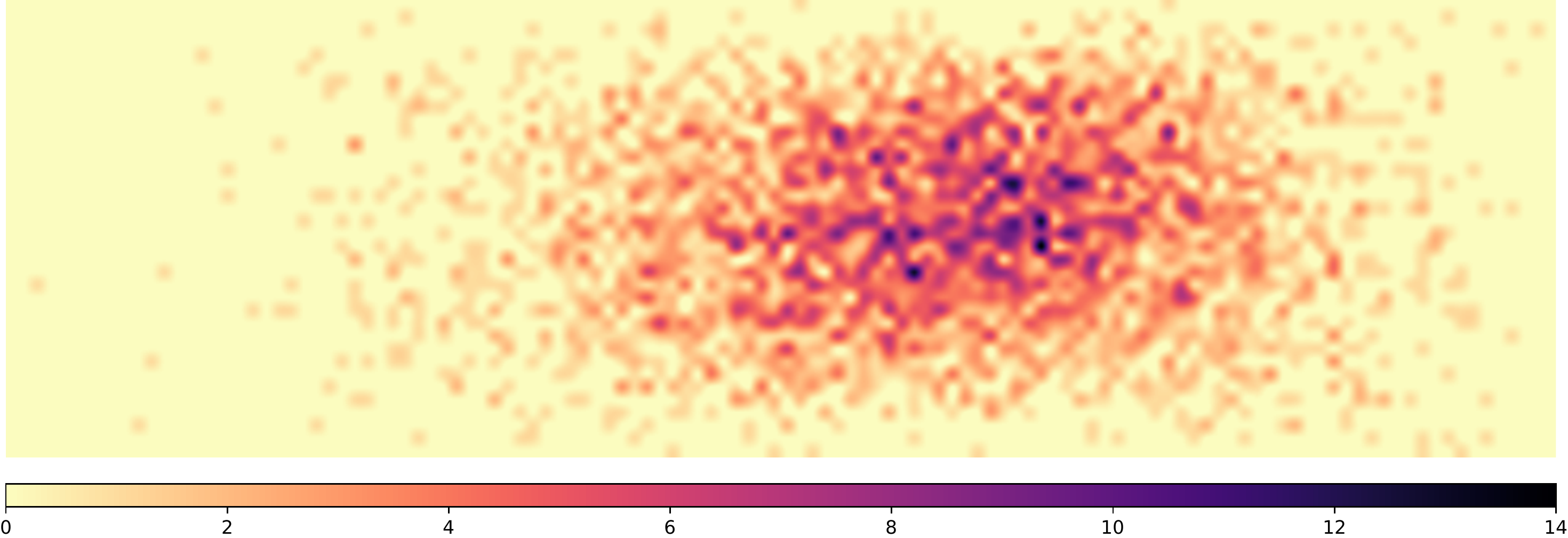} }}
    \subfloat{{\includegraphics[width=4.3cm]{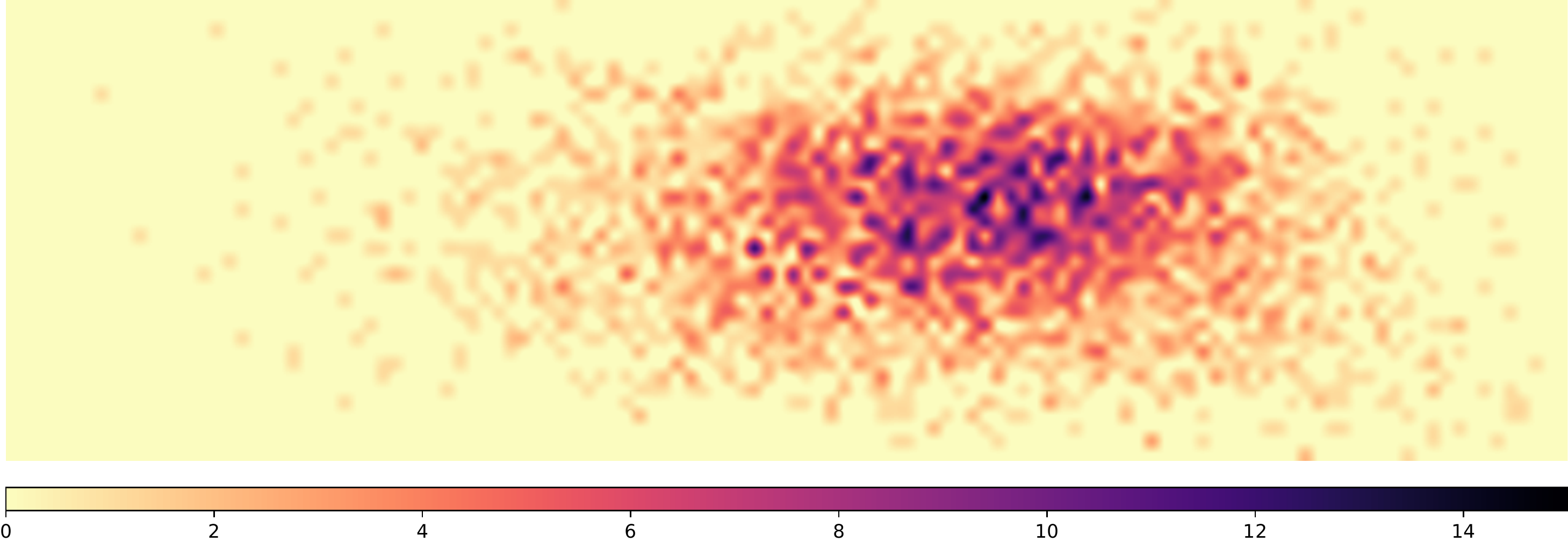} }}
    
    \caption{Heatmap of observations for the entire sequence 2. The first observation is randomly defined, and the RL agent chooses the others based on the learned policy.}
    \label{fig:baseline_heatmap}
    \vspace{-1em}
\end{figure*}

\subsection{Evaluation Metrics}

The most common evaluation metrics for visual odometry compute the agent's absolute trajectory error (ATE) and the relative pose error (RPE). These metrics are commonly reported for the entire trajectory by computing the root mean square error (RMSE) for all frames.

\textbf{Absolute Trajectory Error (ATE)} is used to compute the method's global consistency. We compare the estimated pose with the ground-truth pose for each frame. However, the poses usually are specified in arbitrary coordinate frames and must be aligned to be compared. The absolute pose error at instant $i$ is given by $\mathcal{E}_i = \textbf{G}_i^{-1}\textbf{A}\textbf{H}_i$, where $\textbf{G}_i$ is the ground-truth pose at instant $i$, $\textbf{H}_i$ is the estimated trajectory pose at instant $i$, $\textbf{A}$ is the best alignment transformation.

\textbf{Relative Pose Error (RPE)} computes the error using only a relative relationship between frames, which could solve the issues associated with a global frame comparison. In this sense, RPE measures the local consistency of the trajectory and is a reliable metric for the drift. The relative pose error at instant $i$ is given by $   \mathcal{F}_i = \frac{\textbf{H}_i^{-1} \textbf{H}_{i+k}}{\textbf{G}_i^{-1} \textbf{G}_{i+k}}$, where $k$ is a fixed time interval, which determines the trajectory consistency accuracy; for visual odometry, $k=1$ is usually used. We compute RPE by averaging all sub-sequences ranging from 100 to 800 meters in KITTI's sequences.

\subsection{Hardware and Hyperparameters}

We implemented this work in Python 3 with Pytorch. The hardware used for building and training the models was an Intel Core i7-10700KF @ 3.80GHz, Nvidia RTX 2060 with 6Gb, and Cuda v11.1. The model configuration consisted of 3 image patches of $32 \times 32$ pixels, batch size of 128, supervised learning rate of \num{1e-4}, and RL learning rate of \num{1e-6}; we employed the Adam optimizer for both networks. We trained the models for 400 epochs without early stopping or learning rate decay. The average training time was around 13 hours. The inference time is 35ms for a pair of frames.

\section{Experimental Results}

We conducted several experiments (Table~\ref{table:baseline_configurations}) to provide a better understanding of the RAM-VO behavior on complex sequences in the KITTI dataset~\cite{geiger_vision_2013}. First, we validated the number of glimpses/observations on building a usable internal state $\textbf{h}_t$ for visual odometry. Thus, we varied the number of observations from 1 to 12 and also tested with random observations. Second, we replaced the REINFORCE algorithm with PPO to evaluate the impact of the policy on the generalization; finally, we varied the internal state $\textbf{h}_t$ capacity from 1024 to 256 hidden units to evaluate the impact on the drift error.

\subsection{Number of Observations}

The first experiment consisted of making a single observation at the image's center. The error metrics indicate that the model overfits the train sequences and cannot generalize, possibly by learning the appearance instead of geometry. The information captured from only a central region facilitates the learning of the scale since it tends to be constant for most frames; however, some frames show different behaviors (e.g., obstructions, reflections) and quickly degrades the trajectory prediction. Also, only one capture in the same location reduces the data diversity necessary to learn complex behaviors; the model tends to memorize the single observation instead of learning a general dynamics. 

The second experiment consisted of a single observation in a random location. Although a single random observation provides more data diversity during training, it still captures little and sparse information, affecting the scale learning, hampering robust predictions. Furthermore, good predictions with random observations require learning the general dynamics since the input space is ample and the model's capacity is limited. Finally, we highlight that these experiments did not use reinforcement learning since the location is already determined; and a single observation did not provide enough information for generalization.

The subsequent experiments aim to determine the impact of the observations, and consequently, the learned policy to select informative patches and the core network's ability to integrate them. Therefore, we set the number of observations to 4, 8, and 12; all locations are determined by the policy. The model achieved the best generalization results with 4 and 8 glimpses; 12 glimpses have not provided better results, as more data not necessarily mean better predictions. More observations demand more from the core and locator networks since a single poor observation can harm the entire internal state and delay, even more, the sparse reward. 

Considering the experiment with 8 observations, the totality of input information is 5.7\% of the total available. We conclude that the agent is retrieving high informative patches for most observations, which consisted of edges and corners (Figure~\ref{fig:baseline_glimpses}). Also, the learned policy displayed the traditional Gaussian pattern indicating a preference for observing the right-center portion of the image (Figure~\ref{fig:baseline_heatmap}).

\begin{table}[b!]
\centering
\caption{The impact of the number of observations on the average RPE and ATE for all sequences on train and test set. $\mathrm{\bar{t}_{rpe}}$ represents the average translational RMSE drift (\%) on length of 100m to 800m. $\mathrm{\bar{r}_{rpe}}$ is the average rotational RMSE drift (\degree/100m) on length of 100m to 800m. $\mathrm{\overline{ATE}}$ represents the average absolute trajectory error.}
\label{table:baseline_configurations}

\begin{tabular}{p{2.15cm}p{0.6cm}p{0.5cm}p{0.7cm}p{0.6cm}p{0.5cm}p{0.6cm}}
\toprule
                       & \multicolumn{3}{c}{\textbf{Train set}} & \multicolumn{3}{c}{\textbf{Test set}}                   \\ \cline{2-7} 
\textbf{Configuration} & $\mathrm{\bar{t}_{rpe}}$       & $\mathrm{\bar{r}_{rpe}}$         & $\mathrm{\overline{ATE}}$         & $\mathrm{\bar{t}_{rpe}}$ & $\mathrm{\bar{r}_{rpe}}$ &$\mathrm{\overline{ATE}}$  \\ \hline
1 glimpse at center    &   0.984 & 0.408 & 3.216 & 16.369 &	7.666 & 36.104 \\ \hline
1 glimpse random       & 16.516 &	4.940 &	127.158 &  25.969	& 7.939 &	42.221   \\ \hline
4 glimpses             &  3.599 &	1.608 &	36.463  & 10.929 &	\textbf{3.985} &	\textbf{17.418}   \\ \hline
8 glimpses             & \textbf{3.021} &	\textbf{1.393} &	\textbf{20.311} & \textbf{10.888} &	4.206 &	19.806            \\ \hline
8 glimpses random      &  5.227 &	2.126 &	58.193  &  12.461 &	4.127 &	21.004    \\
\hline
12 glimpses            & 3.335&	1.504 &	32.517 & 13.181 &	5.771 &	24.762  \\  \bottomrule
\end{tabular}
\end{table}

\subsection{Experiments with PPO}

The following experiments (Table~\ref{table:ppo_configurations}) evaluate the results achieved by replacing the REINFORCE algorithm with PPO. We also investigated the impact of reducing the internal state $\textbf{h}_t$ capacity from 1024 to 256 hidden units. The core network corresponds to most of the model's parameters; therefore, knowing the minimum capacity required to achieve good results is crucial for delivering lightweight models. For all experiments, we captured 8 glimpses and computed the statistics for three distinct executions. 

\begin{table}[h!]
\centering

\caption{The impact of PPO and the internal state capacity. For metrics explanation, see Table~\ref{table:baseline_configurations}.}
\label{table:ppo_configurations}

\begin{tabular}{p{0.6cm}p{0.8cm}p{0.6cm}p{0.6cm}p{0.6cm}p{0.6cm}p{0.6cm}p{0.6cm}}
\toprule
 & & \multicolumn{3}{c}{\textbf{Train set}} & \multicolumn{3}{c}{\textbf{Test set}}                  
 \\ \cline{3-8} 
\textbf{Config.} & \textbf{Param.} & $\mathrm{\bar{t}_{rpe}}$       & $\mathrm{\bar{r}_{rpe}}$         & $\mathrm{\overline{ATE}}$         & $\mathrm{\bar{t}_{rpe}}$ & $\mathrm{\bar{r}_{rpe}}$ &$\mathrm{\overline{ATE}}$  \\ \hline
PPO 1024   & \vspace{-0.1em} 16.54M & $\textbf{4.095} \pm 0.27$ &	$\textbf{1.681} \pm 0.17$ & $\textbf{30.066} \pm 9.89$ & $\textbf{10.634}  \pm 1.38$ &	$\textbf{4.369}  \pm  0.43$ &	$\textbf{17.181}  \pm  2.20$\\ \hline

PPO 512    & \vspace{-0.1em} 5.84M &  $7.068	\pm 2.43$ & $2.735 \pm 0.74$ &	$54.587 \pm 26.91$ & $15.486 \pm 4.06$ &	$7.084 \pm 1.87$ &	$24.810 \pm 6.67$ \\ \hline

PPO 256     & \vspace{-0.1em} 2.92M & $7.287 \pm	1.68$ & $2.791 \pm 0.44$ &	$52.211 \pm 21.08$ & $15.499 \pm 5.18$ &	$6.685 \pm 2.00$ &	$25.624 \pm 8.88$ \\  \bottomrule

\end{tabular}

\end{table}

\begin{figure}[tb!]
    \centering
    \subfloat{{\includegraphics[width=4.2cm]{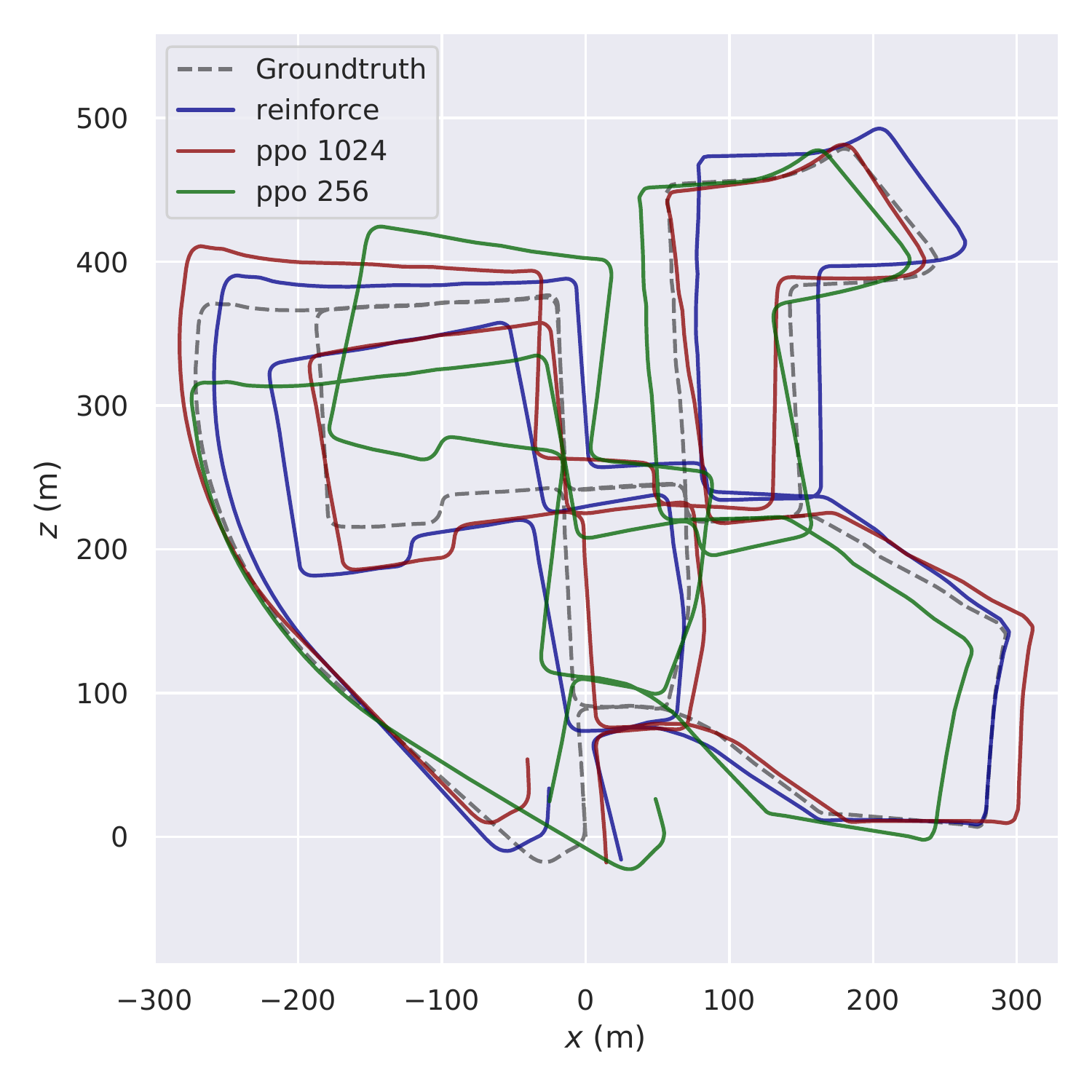} }}
    \subfloat{{\includegraphics[width=4.2cm]{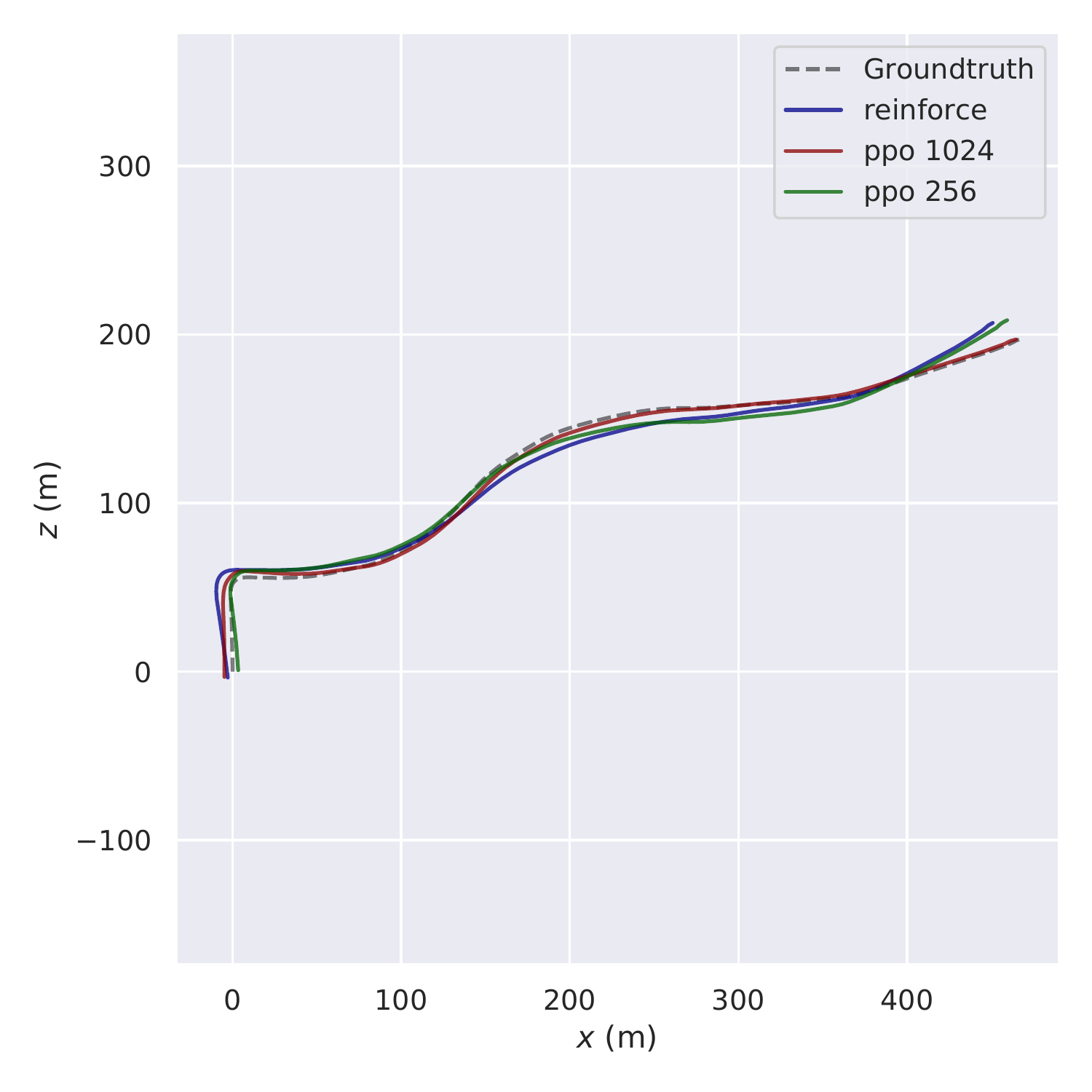} }}
    \vspace{-1.3em}
    \subfloat{{\includegraphics[width=4.2cm]{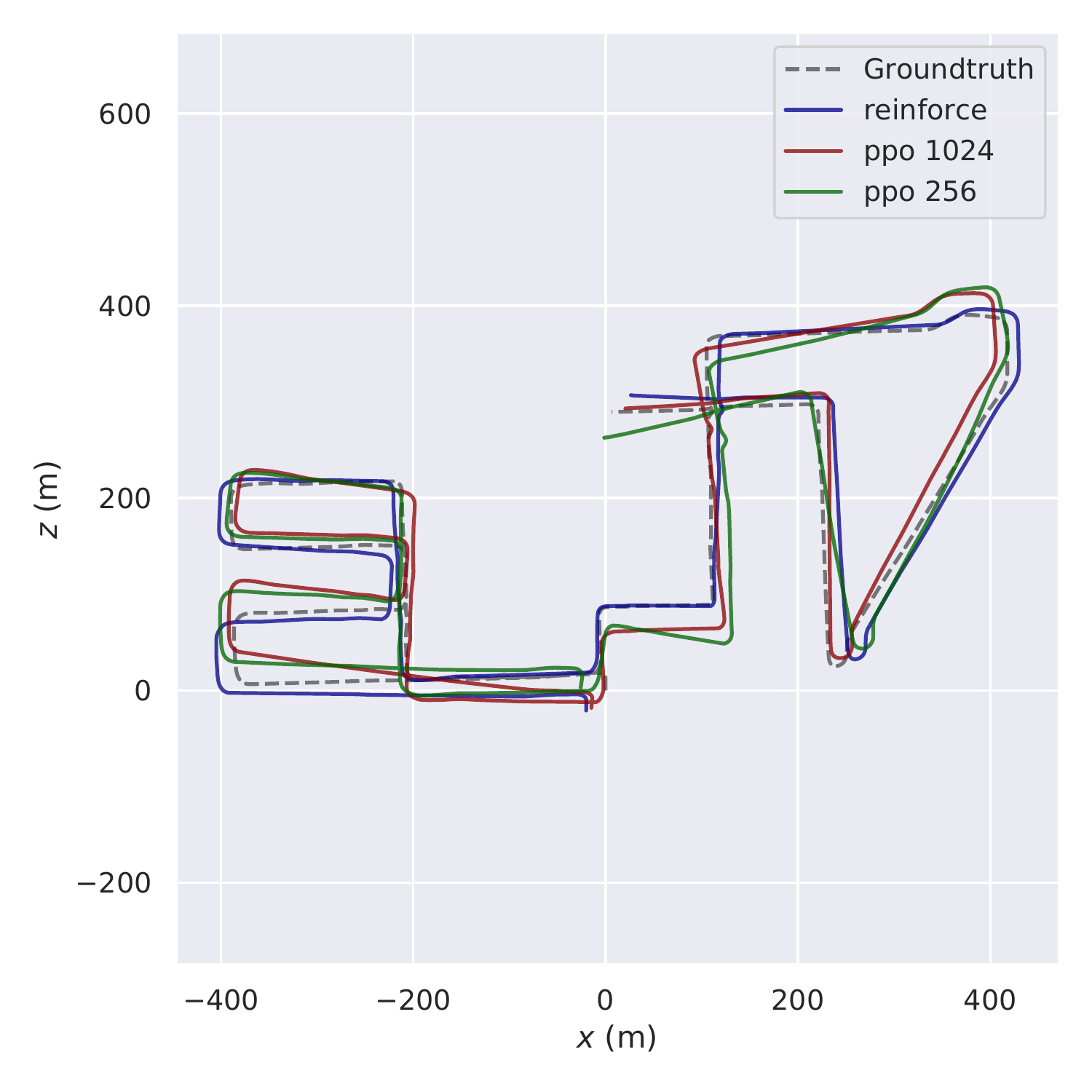} }}
    \subfloat{{\includegraphics[width=4.2cm]{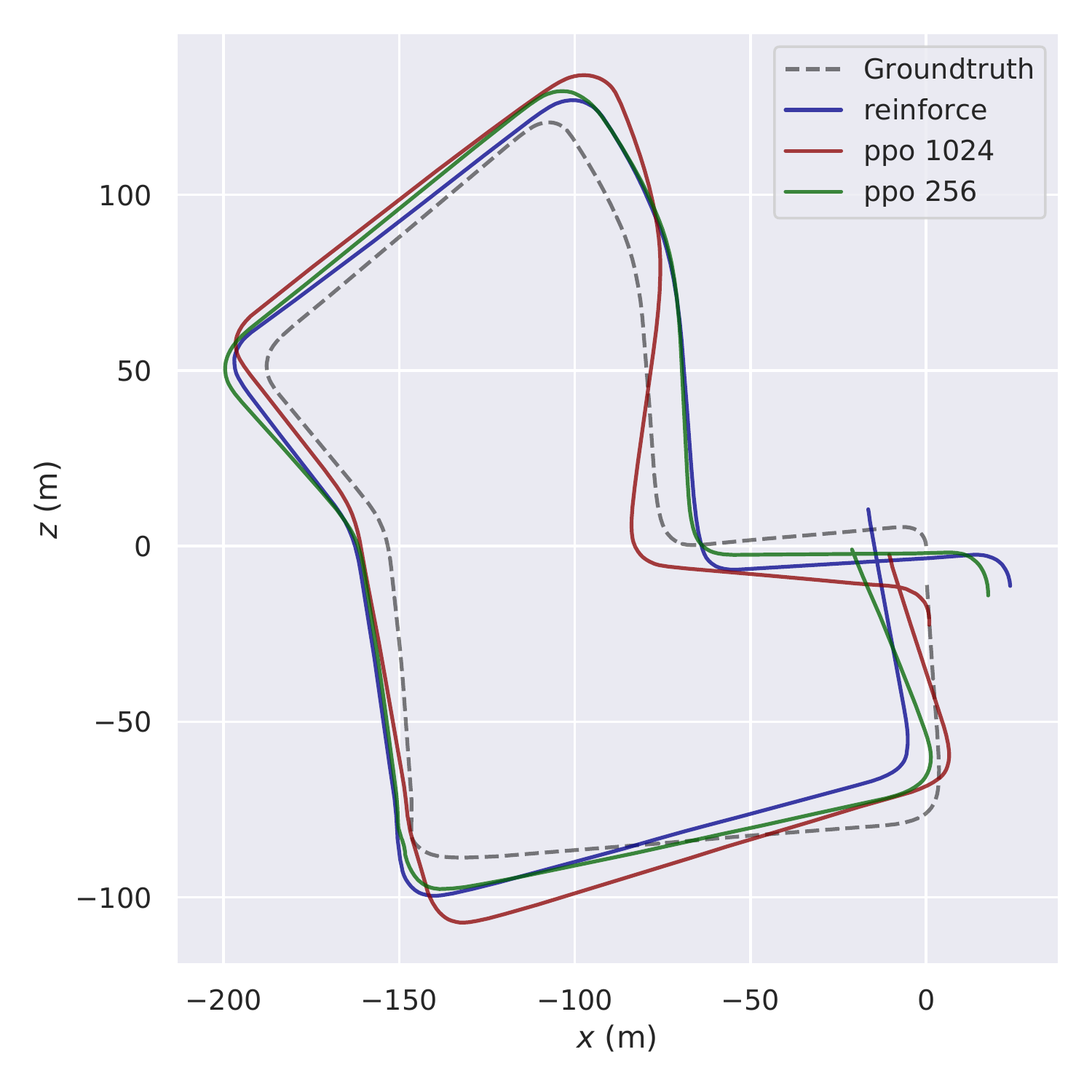} }}
    \vspace{-1.3em}
    \subfloat{{\includegraphics[width=4.2cm]{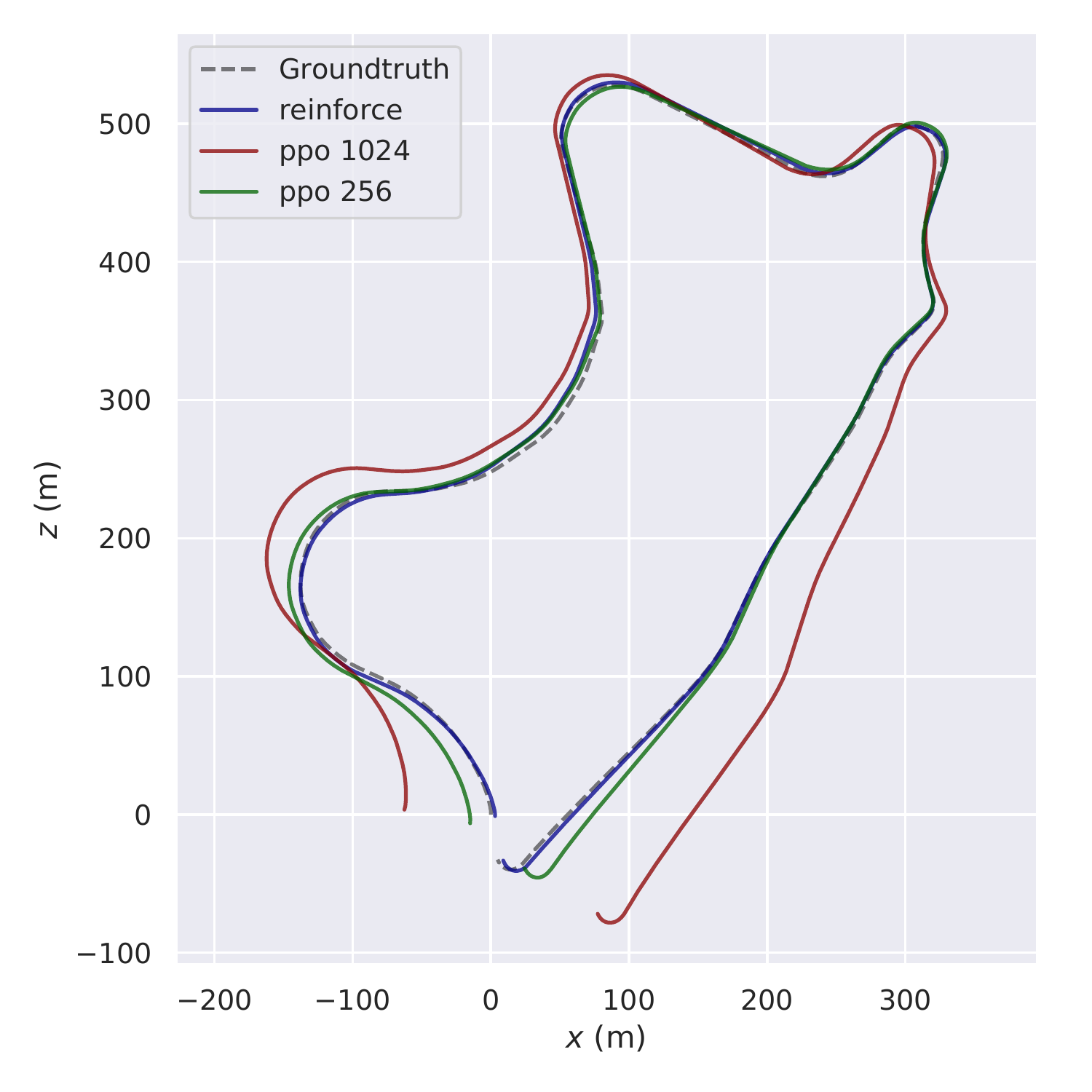} }}
    \subfloat{{\includegraphics[width=4.2cm]{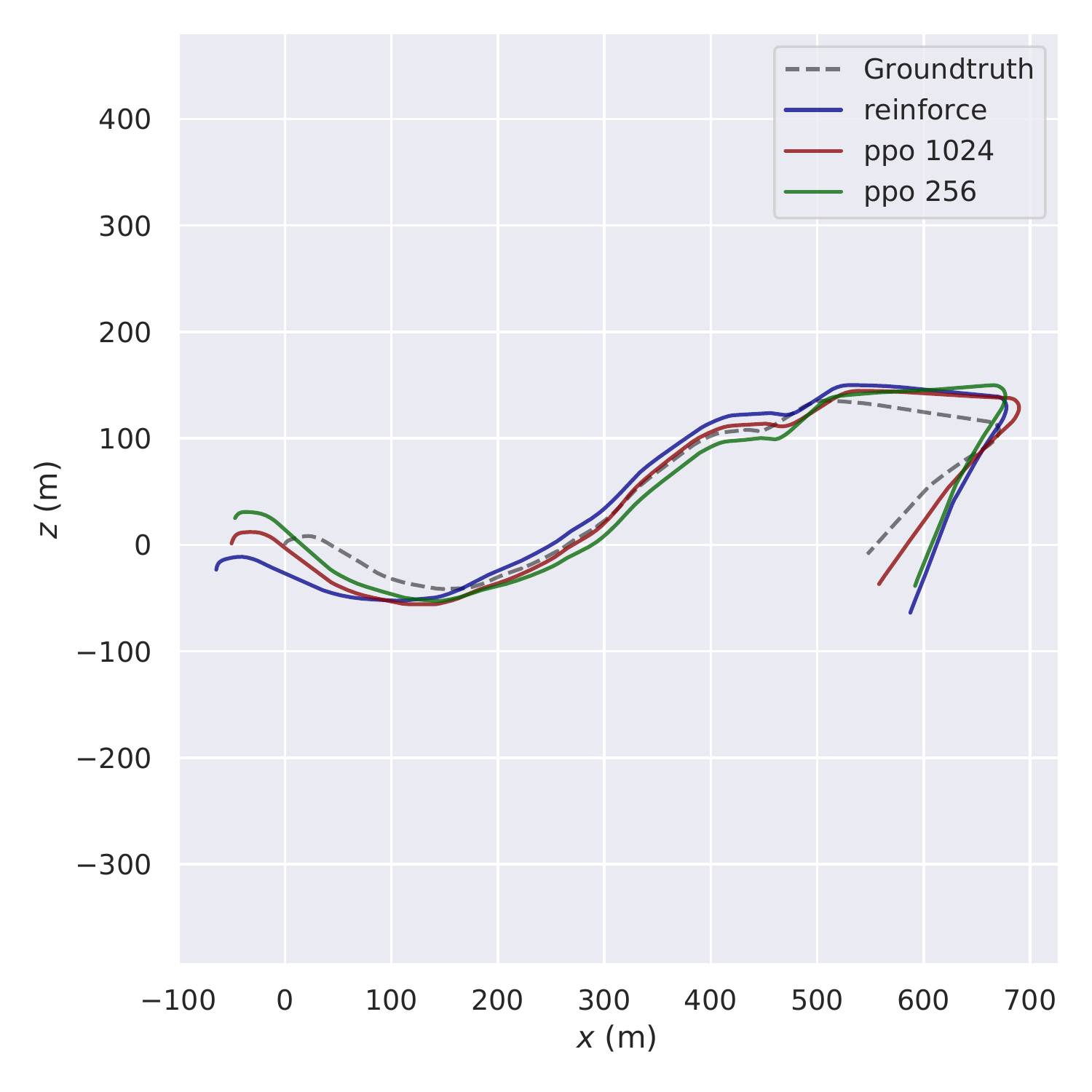} }}
    \vspace{-1em}
    \caption{RAM-VO trajectories' predictions for training (left column) and testing (right column) sequences on the KITTI dataset using our best model.}
    \label{fig:baseline_trajectory}
    \vspace{-1em}
\end{figure}

PPO with 1024 hidden units provided the best generalization, mainly due to the increased capacity. We observed that decreasing the number of parameters increases the relative error during training, and the generalization is strongly affected on average. Although PPO 1024 had the best performance, the difference in the results' quality may not justify a three-fold increase in parameters. PPO algorithm is sensitive to the initialization; therefore, we selected the best models of three executions to predict the trajectories (Figure~\ref{fig:baseline_trajectory}). PPO 256 has only 2.92 million parameters and provides results compatible with the PPO 1024 on the best execution. In conclusion, we asserted that the PPO algorithm provided a slightly better generalization capacity than the REINFORCE algorithm. PPO learned a more centered policy, although very similar to the one learned by the REINFORCE algorithm.

\subsection{Comparison with Literature}

Our best RAM-VO with PPO obtained competitive results (Table~\ref{table:comparison_literature}) using less input information than similar methods, around 5.7\% of the total available, considering 8 glimpses with the size of $32 \times 32$ pixels. While RAM-VO uses top-down attention to capture regions of interest, methods like ORB-SLAM~\cite{mur2015orb} need to analyze an entire image to detect keypoints. Besides, ORB-SLAM is a geometric method that depends on high-texture regions for an accurate keypoint match between frames. DeepVO~\cite{wang_deepvo_2017} and ESP-VO~\cite{wang_end--end_2018} are both based on the FlowNet~\cite{fischer_flownet_2015} architecture and therefore have more convolutional layers and channels than RAM-VO. These architectures perform direct VO  by determining the frames' correspondence from the pixel values; consequently, they are more robust to outliers than ORB-SLAM. However, direct methods are costly, especially DeepVO and ESP-VO, which use the entire image as input. In visual odometry, the motion is present in the whole image; hence, more data does not necessarily bring novel information. 

\begin{table}[b!]
\centering

\caption{RAM-VO results compared to other methods on test sequences. For metrics explanation, see Table~\ref{table:baseline_configurations}.  }
\label{table:comparison_literature}

\begin{tabular}{p{2cm}p{0.4cm}p{0.5cm}p{0.4cm}p{0.4cm}p{0.5cm}p{0.5cm}p{0.5cm}}
\toprule
& \textbf{Data} & \multicolumn{2}{c}{\textbf{Seq. 03}}     
& \multicolumn{2}{c}{\textbf{Seq. 07}}   
& \multicolumn{2}{c}{\textbf{Seq. 10}} 
\\ \cline{3-8} 
\textbf{Method} & \textbf{\%} & $\mathrm{t_{rpe}}$ & $\mathrm{r_{rpe}}$ & $\mathrm{t_{rpe}}$ & $\mathrm{r_{rpe}}$ & $\mathrm{t_{rpe}}$ & $\mathrm{r_{rpe}}$ 
\\ \hline
ORB-SLAM~\cite{mur2015orb} & - & 21.07 & 18.36 & 24.53 & 38.90 & 86.51 & 98.90   \\  \hline
DeepVO~\cite{wang_deepvo_2017} & 100 & 8.49 & 6.89 & 3.91 & 4.60 & \textbf{8.11} & 8.83  \\ \hline
ESP-VO~\cite{wang_end--end_2018} & 100 & 6.72 & 6.46 & \textbf{3.52} & 5.02 & 9.77 & 10.20  \\  \hline
RAM-VO 1024 & \textbf{5.68} & \textbf{5.72} & \textbf{3.08} & 9.17 & 5.63 & 13.85 & \textbf{3.24} \\ \hline
RAM-VO 256 & \textbf{5.68} & 7.08 & 4.01 & 7.55 & \textbf{4.30} & 15.02 & 5.12   \\
\bottomrule
\end{tabular}
\end{table}

The model's capacity is also a relevant fact to be considered. Although DeepVO and ESP-VO may present similar results on average, RAM-VO with 256 hidden units in the core network achieves comparable results with only 2.7 million parameters --- which is much lower than the 17 million parameters reached by RAM-VO with 1024 hidden units. Concurrent methods regularly pass 32 million parameters, especially when they are extensions of architectures like AlexNet~\cite{krizhevsky_imagenet_2012}, and FlowNet~\cite{fischer_flownet_2015}. Also, these CNN networks are considerably deep and add a high cost in terms of trainable parameters, making the training slow and the deployment a complicated task for mobile devices.

Learning-based methods tend to be slow and costly to run due to requiring high-end devices with large processing power, GPUs, and better batteries; these issues make adopting learning methods problematic. In the same way, the results presented by ORB-SLAM are worse compared to learning-based methods, but they are sufficiently fast to build online applications on mobile devices. In this context, RAM-VO represents an alternative method capable of providing results similar to large models but with a smaller cost in trainable parameters and data consumption.

\subsection{Limitations}

RAM-VO has difficulty in obtaining the world scale and determining the translational motion with greater precision. This issue probably happens due to the small image patches that, depending on the vehicle's velocity, can prevent the overlap between frames, compromising the regression since it relies on the features' correspondence. We aim to assess these problems in future formulations.

\section{Conclusion}

In this work, we proposed the RAM-VO model for monocular end-to-end visual odometry. Our model was extended from RAM~\cite{mnih2014recurrent} and therefore implemented attention and reinforcement learning to optimize the selection of visual information for regression tasks. RAM-VO innovated with the addition of 6-DoF pose regression, a robust glimpse network to learn optical flow, an improved core network to store temporal relation between observations, and the replacement of REINFORCE by the PPO algorithm to learn better policies. To the best of our knowledge, RAM-VO is the first architecture for visual odometry that implements reinforcement learning in part of the pipeline. The experimental results indicate that RAM-VO can predict 6-DoF poses in the real-world KITTI~\cite{geiger_vision_2013} dataset with generalization for unseen sequences. The comparison with the literature indicated that RAM-VO could achieve competitive results using significantly less trainable parameters and input information. Similar learning methods consume the whole input image to determine the pose, while the RAM-VO uses a small fraction, around 5.7\% of the total input data.


\section*{Acknowledgment}

The authors would like to thank the Brazilian National Council for Scientific and Technological Development (CNPq), grant 130834/2019-0, and Bradesco Bank.

\bibliographystyle{Transactions-Bibliography/IEEEtran}
\bibliography{bib/datasets.bib, bib/others.bib, bib/old.bib, bib/books.bib, bib/supervised.bib, bib/unsupervised.bib}

\begin{thebibliography}{10}
\providecommand{\url}[1]{#1}
\csname url@samestyle\endcsname
\providecommand{\newblock}{\relax}
\providecommand{\bibinfo}[2]{#2}
\providecommand{\BIBentrySTDinterwordspacing}{\spaceskip=0pt\relax}
\providecommand{\BIBentryALTinterwordstretchfactor}{4}
\providecommand{\BIBentryALTinterwordspacing}{\spaceskip=\fontdimen2\font plus
\BIBentryALTinterwordstretchfactor\fontdimen3\font minus
  \fontdimen4\font\relax}
\providecommand{\BIBforeignlanguage}[2]{{%
\expandafter\ifx\csname l@#1\endcsname\relax
\typeout{** WARNING: IEEEtran.bst: No hyphenation pattern has been}%
\typeout{** loaded for the language `#1'. Using the pattern for}%
\typeout{** the default language instead.}%
\else
\language=\csname l@#1\endcsname
\fi
#2}}
\providecommand{\BIBdecl}{\relax}
\BIBdecl

\bibitem{geiger_vision_2013}
A.~Geiger, P.~Lenz, C.~Stiller, and R.~Urtasun,
  ``\BIBforeignlanguage{en}{Vision meets robotics: {The} {KITTI} dataset},''
  \emph{\BIBforeignlanguage{en}{INT J ROBOT RES}}, vol.~32, no.~11, pp.
  1231--1237, Sep. 2013.

\bibitem{goodfellow_deep_2016}
I.~Goodfellow, Y.~Bengio, and A.~Courville, \emph{\BIBforeignlanguage{en}{Deep
  learning}}, ser. Adaptive computation and machine learning.\hskip 1em plus
  0.5em minus 0.4em\relax Cambridge, Massachusetts: The MIT Press, 2016.

\bibitem{ciarfuglia_evaluation_2014}
T.~A. Ciarfuglia, G.~Costante, P.~Valigi, and E.~Ricci,
  ``\BIBforeignlanguage{en}{Evaluation of non-geometric methods for visual
  odometry},'' \emph{\BIBforeignlanguage{en}{Robotics and Autonomous Systems}},
  vol.~62, no.~12, pp. 1717--1730, 2014.

\bibitem{guizilini_semi-parametric_2013}
V.~Guizilini and F.~Ramos, ``\BIBforeignlanguage{en}{Semi-parametric learning
  for visual odometry},'' \emph{\BIBforeignlanguage{en}{INT J ROBOT RES}},
  vol.~32, no.~5, pp. 526--546, Apr. 2013.

\bibitem{roberts_memory-based_2008}
R.~Roberts, H.~Nguyen, N.~Krishnamurthi, and T.~Balch, ``Memory-based learning
  for visual odometry,'' in \emph{2008 {IEEE ICRA}}, May 2008, pp. 47--52,
  iSSN: 1050-4729.

\bibitem{mnih2014recurrent}
V.~Mnih, N.~Heess, A.~Graves, and K.~Kavukcuoglu, ``Recurrent models of visual
  attention,'' \emph{arXiv preprint arXiv:1406.6247}, 2014.

\bibitem{lecun1998mnist}
Y.~LeCun, ``The mnist database of handwritten digits,'' \emph{http://yann.
  lecun. com/exdb/mnist/}, 1998.

\bibitem{williams1992simple}
R.~J. Williams, ``Simple statistical gradient-following algorithms for
  connectionist reinforcement learning,'' \emph{Machine learning}, vol.~8, no.
  3-4, pp. 229--256, 1992.

\bibitem{schulman2017proximal}
J.~Schulman, F.~Wolski, P.~Dhariwal, A.~Radford, and O.~Klimov, ``Proximal
  policy optimization algorithms,'' \emph{arXiv preprint arXiv:1707.06347},
  2017.

\bibitem{konda_learning_2015}
K.~Konda and R.~Memisevic, ``\BIBforeignlanguage{en}{Learning {Visual}
  {Odometry} with a {Convolutional} {Network}:},'' in
  \emph{\BIBforeignlanguage{en}{Proceedings of the 10th {International}
  {Conference} on {Computer} {Vision} {Theory} and {Applications}}}.\hskip 1em
  plus 0.5em minus 0.4em\relax Berlin, Germany: SCITEPRESS - Science and and
  Technology Publications, 2015, pp. 486--490.

\bibitem{muller_flowdometry_2017}
P.~Muller and A.~Savakis, ``\BIBforeignlanguage{en}{Flowdometry: {An} {Optical}
  {Flow} and {Deep} {Learning} {Based} {Approach} to {Visual} {Odometry}},'' in
  \emph{\BIBforeignlanguage{en}{2017 {IEEE} {WACV}}}.\hskip 1em plus 0.5em
  minus 0.4em\relax Santa Rosa, CA, USA: IEEE, Mar. 2017, pp. 624--631.

\bibitem{mohanty_deepvo_2016}
V.~Mohanty, S.~Agrawal, S.~Datta, A.~Ghosh, V.~D. Sharma, and D.~Chakravarty,
  ``{DeepVO}: {A} {Deep} {Learning} approach for {Monocular} {Visual}
  {Odometry},'' \emph{arXiv:1611.06069 [cs]}, Nov. 2016.

\bibitem{wang_deepvo_2017}
S.~Wang, R.~Clark, H.~Wen, and N.~Trigoni, ``{DeepVO}: {Towards} {End}-to-{End}
  {Visual} {Odometry} with {Deep} {Recurrent} {Convolutional} {Neural}
  {Networks},'' \emph{2017 IEEE ICRA}, pp. 2043--2050, May 2017.

\bibitem{peretroukhin_dpc-net_2018}
V.~Peretroukhin and J.~Kelly, ``\BIBforeignlanguage{en}{{DPC}-{Net}: {Deep}
  {Pose} {Correction} for {Visual} {Localization}},''
  \emph{\BIBforeignlanguage{en}{IEEE Robotics and Automation Letters}}, vol.~3,
  no.~3, pp. 2424--2431, Jul. 2018.

\bibitem{zhao_learning_2018}
C.~Zhao, L.~Sun, P.~Purkait, T.~Duckett, and R.~Stolkin, ``Learning monocular
  visual odometry with dense {3D} mapping from dense {3D} flow,''
  \emph{arXiv:1803.02286 [cs]}, Jul. 2018, arXiv: 1803.02286.

\bibitem{valada_deep_2018}
A.~Valada, N.~Radwan, and W.~Burgard, ``Deep {Auxiliary} {Learning} for
  {Visual} {Localization} and {Odometry},'' \emph{arXiv:1803.03642 [cs]}, Mar.
  2018, arXiv: 1803.03642.

\bibitem{saputra_distilling_2019}
M.~R.~U. Saputra, P.~P.~B. de~Gusmao, Y.~Almalioglu, A.~Markham, and
  N.~Trigoni, ``Distilling {Knowledge} {From} a {Deep} {Pose} {Regressor}
  {Network},'' \emph{arXiv:1908.00858 [cs]}, Aug. 2019.

\bibitem{saputra_learning_2019}
M.~R.~U. Saputra, P.~P.~B. de~Gusmao, S.~Wang, A.~Markham, and N.~Trigoni,
  ``Learning {Monocular} {Visual} {Odometry} through {Geometry}-{Aware}
  {Curriculum} {Learning},'' \emph{arXiv:1903.10543 [cs]}, Nov. 2019.

\bibitem{damirchi2020exploring}
H.~Damirchi, R.~Khorrambakht, and H.~D. Taghirad, ``Exploring self-attention
  for visual odometry,'' \emph{arXiv preprint arXiv:2011.08634}, 2020.

\bibitem{kuo2020dynamic}
X.-Y. Kuo, C.~Liu, K.-C. Lin, and C.-Y. Lee, ``Dynamic attention-based visual
  odometry,'' in \emph{Proceedings of the IEEE/CVF CVPR Workshops}, 2020, pp.
  36--37.

\bibitem{liang_salientdso_2018}
H.-J. Liang, N.~J. Sanket, C.~Fermüller, and Y.~Aloimonos, ``{SalientDSO}:
  {Bringing} {Attention} to {Direct} {Sparse} {Odometry},''
  \emph{arXiv:1803.00127 [cs]}, Feb. 2018, arXiv: 1803.00127.

\bibitem{engel_direct_2018}
J.~Engel, V.~Koltun, and D.~Cremers, ``\BIBforeignlanguage{en}{Direct {Sparse}
  {Odometry}},'' \emph{\BIBforeignlanguage{en}{IEEE Transactions on Pattern
  Analysis and Machine Intelligence}}, vol.~40, no.~3, pp. 611--625, Mar. 2018.

\bibitem{chen_salient_2019}
M.-y. Chen, C.-l. Wang, and H.-j. Liu, ``\BIBforeignlanguage{en}{Salient
  {FlowNet} and {Decoupled} {LSTM} {Network} for {Robust} {Visual}
  {Odometry}},'' in \emph{\BIBforeignlanguage{en}{2019 {IEEE} {International}
  {Conference} on {Robotics} and {Biomimetics} ({ROBIO})}}.\hskip 1em plus
  0.5em minus 0.4em\relax Dali, China: IEEE, Dec. 2019, pp. 2699--2706.

\bibitem{correia2021attention}
A.~d.~S. Correia and E.~L. Colombini, ``Attention, please! a survey of neural
  attention models in deep learning,'' \emph{arXiv preprint arXiv:2103.16775},
  2021.

\bibitem{fischer_flownet_2015}
A.~Dosovitskiy, P.~Fischer, E.~Ilg, P.~Hausser, C.~Hazirbas, V.~Golkov,
  P.~Smagt, D.~Cremers, and T.~Brox, ``Flownet: Learning optical flow with
  convolutional networks,'' in \emph{2015 IEEE ICCV}.\hskip 1em plus 0.5em
  minus 0.4em\relax Los Alamitos, CA, USA: IEEE Computer Society, dec 2015, pp.
  2758--2766.

\bibitem{mur2015orb}
R.~Mur-Artal, J.~M.~M. Montiel, and J.~D. Tardos, ``Orb-slam: a versatile and
  accurate monocular slam system,'' \emph{IEEE transactions on robotics},
  vol.~31, no.~5, pp. 1147--1163, 2015.

\bibitem{wang_end--end_2018}
S.~Wang, R.~Clark, H.~Wen, and N.~Trigoni,
  ``\BIBforeignlanguage{en}{End-to-end, sequence-to-sequence probabilistic
  visual odometry through deep neural networks},''
  \emph{\BIBforeignlanguage{en}{INT J ROBOT RES}}, vol.~37, no. 4-5, pp.
  513--542, Apr. 2018.

\bibitem{krizhevsky_imagenet_2012}
A.~Krizhevsky, I.~Sutskever, and G.~E. Hinton, ``{ImageNet} {Classification}
  with {Deep} {Convolutional} {Neural} {Networks},'' in \emph{Advances in
  {Neural} {Information} {Processing} {Systems} 25}, F.~Pereira, C.~J.~C.
  Burges, L.~Bottou, and K.~Q. Weinberger, Eds.\hskip 1em plus 0.5em minus
  0.4em\relax Curran Associates, Inc., 2012, pp. 1097--1105.

\end{thebibliography}

\end{document}